\documentclass[10pt,twocolumn,letterpaper]{article}

\usepackage{iccv}
\usepackage{times}
\usepackage{epsfig}
\usepackage{graphicx}
\usepackage{amsmath}
\usepackage{amssymb}
\usepackage{subfigure}
\usepackage{tabularx}
\usepackage{color}
\usepackage{multirow}
\usepackage{colortbl}
\usepackage{epstopdf}
\usepackage{enumitem}

\newcommand{\figref}[1]{Fig. \ref{#1}}
\newcommand{\tabref}[1]{Table \ref{#1}}
\newcommand{\equref}[1]{(\ref{#1})}

\makeatletter
\def\hlinewd#1{%
	\noalign{\ifnum0=`}\fi\hrule \@height #1 \futurelet
	\reserved@a\@xhline}

\makeatletter
\def\thanks#1{\protected@xdef\@thanks{\@thanks
		\protect\footnotetext{#1}}}
\makeatother

\definecolor{srcolor}{rgb}{0,0,0}
\newcommand{\sr}[1]{\textcolor{srcolor}{{#1}}}

\usepackage[breaklinks=true,bookmarks=false]{hyperref}

\iccvfinalcopy 

\def\iccvPaperID{6978} 

\ificcvfinal\fi
\begin{document}
	
\title{Joint Learning of Semantic Alignment and Object Landmark Detection\thanks{This research was supported by R\&D program for Advanced Integrated-intelligence for Identification (AIID) through the National Research Foundation of KOREA (NRF) funded by Ministry of Science and ICT (NRF-2018M3E3A1057289).}}

\author{
	Sangryul Jeon$^1$, Dongbo Min$^2$, Seungryong Kim$^{3}$, Kwanghoon Sohn$^{1,*}$\thanks{$^{*}$Corresponding author}\\
	$^1$Yonsei University, 
	$^2$Ewha Womans University, 
	$^3${\'E}cole Polytechnique F{\'e}d{\'e}rale de Lausanne (EPFL)\\
	\texttt{\{cheonjsr,khsohn\}@yonsei.ac.kr}\\
	\texttt{dbmin@ewha.ac.kr},\texttt{seungryong.kim@epfl.ch}\\
} 

\maketitle
\ificcvfinal\thispagestyle{empty}\fi

\begin{abstract}
	Convolutional neural networks (CNNs) based approaches for semantic alignment and object landmark detection have improved their performance significantly. Current efforts for the two tasks focus on addressing the lack of massive training data through weakly- or unsupervised learning frameworks. In this paper, we present a joint learning approach for obtaining dense correspondences and discovering object landmarks from semantically similar images. Based on the key insight that the two tasks can mutually provide supervisions to each other, our networks accomplish this through a joint loss function that alternatively imposes a consistency constraint between the two tasks, thereby boosting the performance and addressing the lack of training data in a principled manner. To the best of our knowledge, this is the first attempt to address the lack of training data for the two tasks through the joint learning. To further improve the robustness of our framework, we introduce a probabilistic learning formulation that allows only reliable matches to be used in the joint learning process. With the proposed method, state-of-the-art performance is attained on several standard benchmarks for semantic matching and landmark detection, including a newly introduced dataset, JLAD, which contains larger number of challenging image pairs than existing datasets.
\end{abstract}
	
	\section{Introduction}
	Establishing dense correspondences and discovering object landmarks over \textit{semantically} similar images can facilitate a variety of computer vision and computational photography applications~\cite{Hacohen2011,Liu11,novotny18self,zhang18,felzenszwalb2010object}.
	Both tasks aim to understand the underlying structure of an object that is geometrically consistent across different but semantically related instances.
	
	Recently, numerous approaches for the semantic alignment~\cite{rocco17,rocco18,rocco18nips,jeon18,paul18,kim18nips} and object landmark detection~\cite{thewlis17iccv,thewlis17nips,zhang18,jakab18} have been proposed to tackle each problem with deep convolutional neural networks (CNNs) in an end-to-end manner.
	However, supervised training for such tasks often involves in constructing large-scale and diverse annotations of dense semantic correspondence maps or object landmarks.
	Collecting such annotations under large intra-class appearance and shape variations requires a great deal of manual works and is prone to error due to its subjectiveness.
	Consequently, current efforts have focused on using additional constraints or assumptions that help their networks to automatically learn each task in a weakly- or unsupervised setting.
	
	\begin{figure}[t!]
		\centering
		\renewcommand{\thesubfigure}{}
		\subfigure[]
		{\includegraphics[width=1\linewidth]{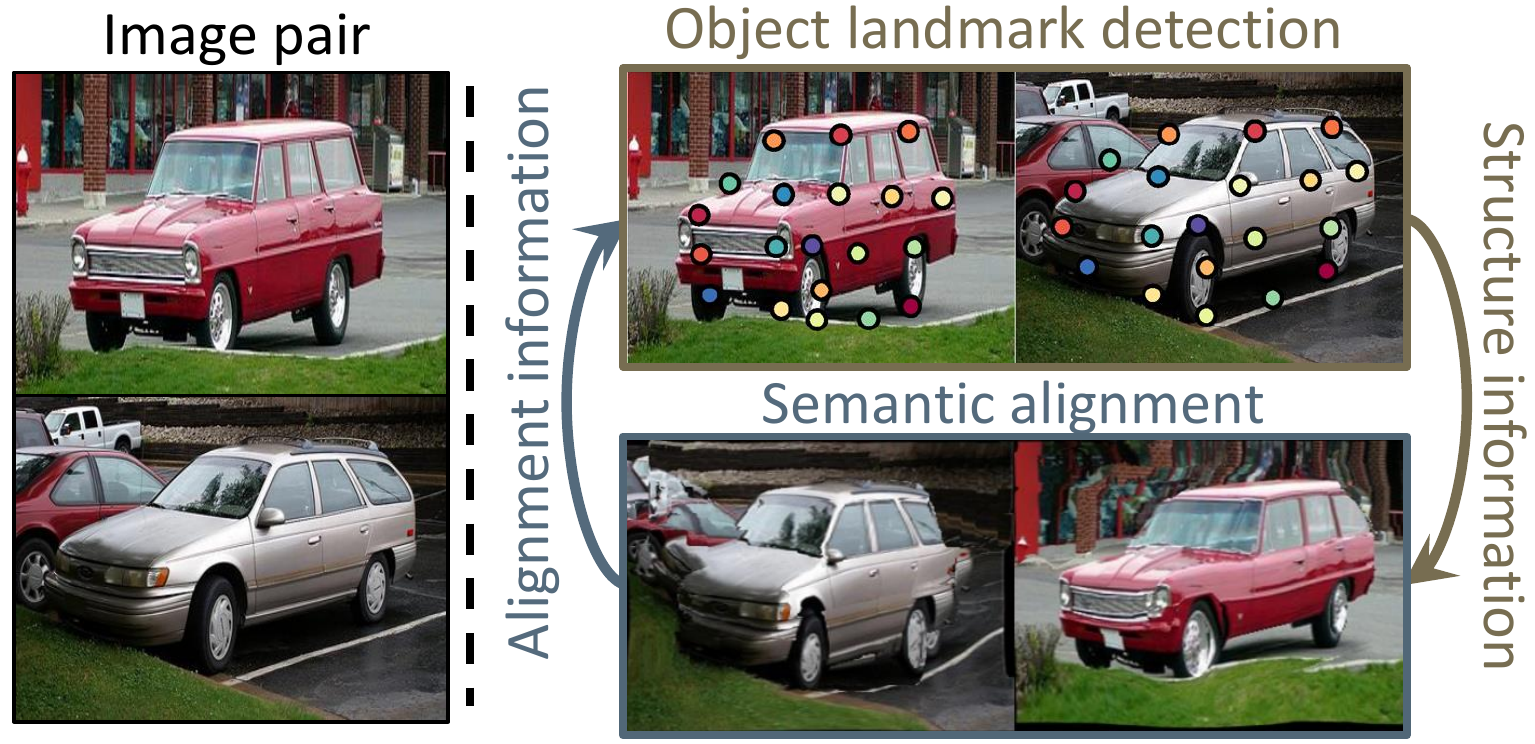}}\hfill
		\vspace{-10pt}	
		\caption{{Illustration of the proposed joint learning framework: given only semantically similar image pairs, we address the crucial drawbacks of current weakly- or unsupervised models for the object landmark detection and semantic alignment task by alternatively leveraging mutual guidance information between them.}}
		\label{img:1}\vspace{-15pt}
	\end{figure}
	
	To overcome the limitations of insufficient training data for semantic correspondence, several works~\cite{rocco17,paul18} have been proposed to utilize 
	a set of sparse corresponding points between source and target images as an additional cue for supervising their networks.
	The key idea is to regulate the densely estimated transformation fields to be consistent with the given sparse corresponding points.
	A possible approach is to synthetically generate the corresponding points from an image itself,
	i.e., by uniformly sampling grid points from a source image and then globally deforming them with random transformations~\cite{kanazawa2016warpnet}.
	However, these synthetic supervisions do not consider photometric variations at all and have difficulties in capturing realisitic geometric deformations.
	Alternatively, several methods~\cite{rocco18,jeon18} alleviate this issue by collecting tentative correspondence samples from real image pairs during training, but this is done in a simple manner, e.g., by thresholding~\cite{rocco18} or checking consistency~\cite{jeon18} with the matching scores.
	More recently, instead of using sparsely collected samples, some methods~\cite{kim18nips,rocco18nips} have employed a complete set of dense pixel-wise matches to estimate locally-varying transformation fields, outperforming previous methods based on a global transfomation model~\cite{rocco17,rocco18,paul18}. However, they often show limited performances in handling relatively large geometric variations due to their weak implicit smoothness constraints such as constraining transformation candidates within local window~\cite{kim18nips} and analyzing local neighbourhood patterns~\cite{rocco18nips}.
	
    Meanwhile, to automatically discover object landmarks without the need of ground-truth labels, following a pioneering work of Thewlis~\etal~\cite{thewlis17iccv}, dense correspondence information across the different instances have been used to impose the equivariance constraint, such that the landmarks should be consistently detectable with respect to given spatial deformations~\cite{thewlis17iccv,thewlis17nips,zhang18,suwajanakorn2018discovery}.
	However, while semantically meaningful and highly accurate correspondences are required to meet the full equivariance, existing techniques mostly rely on synthetic supervisions in a way of generating dense correspondence maps with randomly transformed imagery.
	Similar to existing semantic alignment approaches that leverage synthetic supervision~\cite{rocco17,paul18}, as shown in~\cite{rocco18,jeon18}, they usually do not generalize well to real image pairs and often fail to detect landmarks at semantically meaningful locations of the object.
	
	{In this paper, we present a method for jointly learning object landmark detection and semantic alignment to address the aforementioned limitations of current weakly- or unsupervised learning models of each task.
	As illustrated in~\figref{img:1}, our key observation is that the two tasks are mutually complementary to each other since more realistic and informative supervisions can be provided from their counterparts.
	To be specific, the detected landmarks can offer structure information of an object for the semantic alignment networks where  the estimated correspondence fields are encouraged to be consistent with provided object structures.
	At the same time, densely estimated correspondences across semantically similar image pairs facilitate the landmarks to be consistently localized even under large intra-class variations.
    Our networks accomplish this by introducing a novel joint objective function that alternatively imposes the consistency constraints between the two tasks, thereby boosting the performance and addressing the lack of training data in a principled manner.
	We further improve the robustness of our framework by allowing only reliable matches to be used in the joint learning process through a probabilistic learning formulation of the semantic alignment networks.
	Experimental results on various benchmarks demonstrate the effectiveness of the proposed model over the latest methods for object landmark detection and semantic alignment.}
	
	\section{Related Work}
	
	\paragraph{Semantic alignment}
	
	Recent state-of-the-art techniques for semantic alignment generally regress the transformation parameters directly through an end-to-end CNN model~\cite{rocco17,rocco18,jeon18,kim18nips,paul18}, outperforming conventional methods based on hand-crafted descriptor or optimization~\cite{DSP,Liu11,Ham16}.
	Rocco \etal~\cite{rocco17,rocco18} proposed a CNN architecture that estimates image-level transformation parameters mimicking traditional matching pipeline, such that feature extraction, matching, and parameter regression.
	Seo \etal~\cite{paul18} extended this idea with an offset-aware correlation kernel to focus on reliable correlations, filtering out distractors.
	While providing the robustness against semantic variations to some extent, they have difficulties in yielding fine-grained localization due to the assumption of a global transformation model.
	To address this issue, Jeon \etal~\cite{jeon18} proposed a pyramidal graph model that estimates locally-varying geometric fields with coarse-to-fine scheme.
	Kim \etal~\cite{kim18nips} presented recurrent transformation networks that iteratively align features of source and target and finally obtain precisely refined local translational fields. Rocco \etal~\cite{rocco18nips} proposed to analyze neighbourhood consensus patterns by imposing local constraints to find reliable matches among correspondence candidates.
	However, they rely on weak implicit smoothness constraints such as coarse-to-fine inference~\cite{jeon18}, constrained local search spaces~\cite{kim18nips}, and local neighbourhood consensus~\cite{rocco18nips}.
	In contrast, we explicitly regularize the estimated transformation fields to be consistent with the detected object landmarks through the joint learning process.
    \vspace{-10pt}
    
    \begin{figure*}
    	\centering
    	\renewcommand{\thesubfigure}{}
    	\subfigure[(a)]
    	{\includegraphics[width=0.31\linewidth]{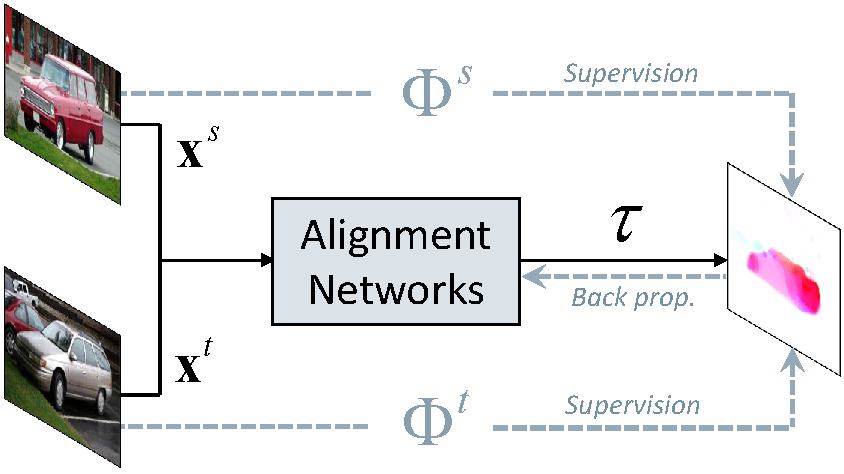}}\quad
    	\subfigure[(b)]
    	{\includegraphics[width=0.31\linewidth]{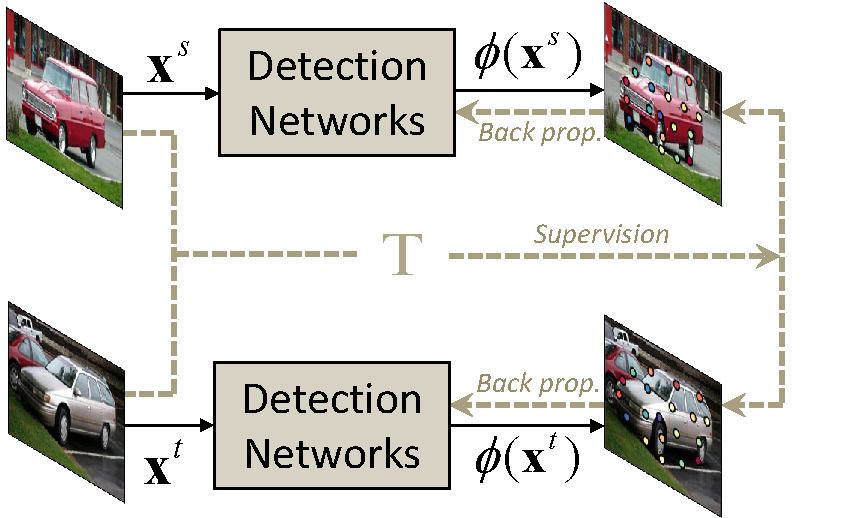}}\quad
    	\subfigure[(c)]
    	{\includegraphics[width=0.31\linewidth]{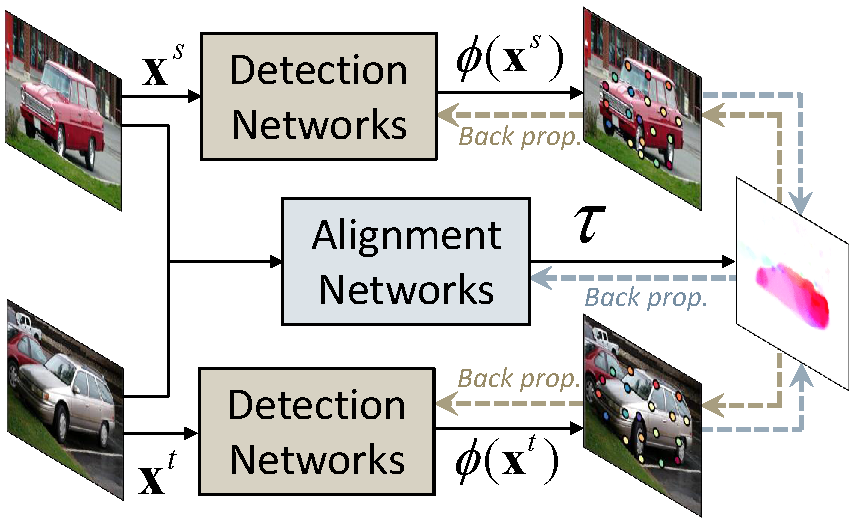}}\quad
    	\vspace{-1pt}
    	\caption{Summary of the methods for: (a) semantic alignment~\cite{rocco17,rocco18,paul18,jeon18}, (b) object landmark detection~\cite{thewlis17iccv,thewlis17nips,zhang18}, and (c) the proposed joint learning framework. Our key observation is that each task can provide an essential supervisory signals to another one. With this motivation, we seamlessly weave both techniques to overcome the lack of training data.}\label{img:2}\vspace{-10pt}
    \end{figure*}
    
	\paragraph{Object landmark detection}
	Methods for unsupervised landmark detection generally rely on the equivariance property such that the object landmarks should be consistently detected with respect to given image deformations.
	As a pioneering work, Thewlis \etal~\cite{thewlis17iccv} proposed to randomly synthesize the image transformations for learning to discover the object landmarks that are equivariant with respect to those transformations.
	They further extended this idea to learn dense object-centric coordinate frame~\cite{thewlis17nips}.
	Both of them rely on the synthetically generated supervisory signals and thus provide inherently limited performance when substantial intra-class variations are given.
	
	Afterward, several works~\cite{zhang18,jakab18} proposed an autoencoding formulation to discover landmarks as explicit structural representations in a way of generating new images conditioned on them. Zhang \etal~\cite{zhang18} proposed to take object landmarks as an intermediate learnable latent variable for reproducing the input image. Jakab \etal~\cite{jakab18} proposed to generate images combining the appearance of the source image and the geometry of the target one by minimizing the perceptual distance.
	However, the ablation studies reported in~\cite{zhang18,jakab18} show that they still rely on the supervision from an image itself such as synthesized image pairs or adjacent frames in videos instead of considering rich appearance variation between different object instances, thus yielding limited performance.
	\vspace{-0pt}
	
	\section{Method}\label{sec:3}
	\subsection{\sr{Problem Statement and Overview}}
	\sr{
        Let us denote \emph{semantically} similar source and target images as $\mathbf{x}^s$ and $\mathbf{x}^t\in \mathbb{R}^{H \times W \times 3}$ where $H$ and $W$ denotes height and width of an image.
		We are interested in learning two mapping functions, $\phi: \mathbf{x} \to \mathbb{R}^{K \times 2}$ that extracts the spatial coordinates of $K$ keypoints from an image $\mathbf{x}$ and $\tau : (\mathbf{x}^s,\mathbf{x}^t) \to \mathbb{R}^{ H \times W \times 2}$ that infers a dense correspondence field from source to target image defined for each pixel in $\mathbf{x}^s$. We specifically learn the two functions through the joint prediction model using only weak supervision in the form of semantically similar image pairs.}
	\sr{		
		To address the insufficient training data for semantic correspondence, several methods~\cite{rocco17,rocco18,paul18,jeon18} utilized a set of sparse corresponding points on the source and target images, called anchor pairs, as an additional cue for supervising their networks.
		The key intuition is that the networks automatically learn to estimate geometric transformation fields over a set of transformation candidates by minimizing the discrepancy between given sparse correspondences.
		Specifically, denoting anchor pairs on source and target image as $\Phi^s$ and $\Phi^t$, they define the semantic alignment loss as
		\begin{equation}\label{equ:2}
			\mathcal{L}_\mathrm{A}(\tau)=\sum\nolimits_n||\Phi_n^t-\tau(\mathbf{x}^s,\mathbf{x}^t) \circ \Phi_n^s||^2,
		\end{equation}
		where $n$ is the number of anchor pairs and $\circ$ is an warping operation.
		This is illustrated in \figref{img:2}(a).}
	\sr{		
		Meanwhile, to address the lack of training data for the landmark detection, the state-of-the-art techniques~\cite{thewlis17nips,thewlis17iccv,zhang18} generally employ dense correspondences between the training image pairs.
		The main idea lies in the equivariance constraint such that the detected landmarks should be equivariant with respect to given geometric deformation.
		Formally, denoting a dense correspondence map between source and target images as $\mathrm{T}$, they aim to learn the landmark detection networks through a siamese configuration by minimizing
		\begin{equation}\label{equ:1}
			\mathcal{L}_\mathrm{D}(\phi)=\sum\nolimits_m||\phi_m(\mathbf{x}^t)-\mathrm{T} \circ \phi_m(\mathbf{x}^s)||^2,
		\end{equation}
		where $m$ is the number of detected landmarks.
		This is illustrated in~\figref{img:2}(b).}

	\begin{figure*}
		\centering
		\renewcommand{\thesubfigure}{}
		\subfigure[]
		{\includegraphics[width=0.8\linewidth]{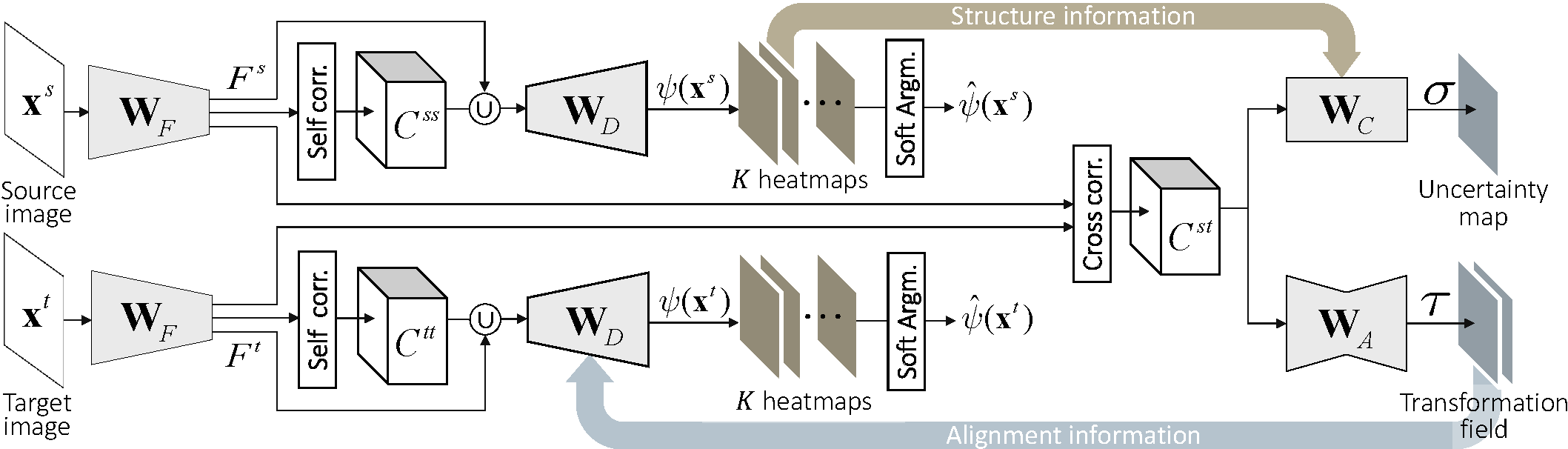}}
		\vspace{-12pt}
		\caption{Network configuration of our framework, consisting of feature extraction networks, landmark detection networks, semantic alignment networks. We alternatively leverage the outputs from each landmark detection and semantic alignment networks as a guidance information for supervising the another one.}\label{img:3}\vspace{-10pt}
	\end{figure*}	
\sr{	
	However, current weakly- or unsupervised learning models for both tasks still suffer from the lack of supervisions of good quality, which may not fully satisfy their consistency constraints.
	To overcome this, we propose to leverage guidance information from each task for supervising another networks, as exemplified in~\figref{img:2}(c).
	The proposed method offers a principled solution that overcomes the lack of massive training data by jointly learning the object landmark detection and semantic alignment in an end-to-end and boosting manner.
	To this end, we introduce a novel joint loss function that alternatively imposes the consistency constraints between the two tasks.
	To further enhance the joint learning process, we propose a probabilistic formulation that predicts and penalizes unreliable matches in the semantic alignment networks.}
	
	\subsection{Network Architectures}
	The proposed networks consist of three sub-networks, including
	feature extraction networks with parameters $\mathbf{W}_F$ to extract feature maps from input images,
	landmark detection networks with parameters $\mathbf{W}_D$ to detect probability maps of landmarks,
	and semantic alignment networks with parameters $\mathbf{W}_A$ and $\mathbf{W}_C$ to infer a geometric transformation field and a uncertainty map, as illustrated in~\figref{img:3}.
	\vspace{-10pt}
	
	\paragraph{Feature extraction and similarity score computation}
	To extract convolutional feature maps of source and target images, the input images are passed through a fully-convolutional feature extraction networks with shared parameters $\mathbf{W}_F$ such that $F = \mathcal{F}(\mathbf{x};\mathbf{W}_F) \in \mathbb{R}^{H \times W \times C}$.
	We share the parameters $\mathbf{W}_F$ for both feature extractions.
	After extracting the features, we normalize them using $L_2$ norm along the $C$ channels.
	
	The similarity
	between two extracted features is then computed as the cosine similarity with $L_2$ normalization:
	\begin{equation}\label{equ:3}
		{C}^{AB}_i = {<F^{A}_i,F^{B}_j>}/{\sqrt{{\sum\nolimits_{l} {<F^A_i,F^B_l>^2}}}},
	\end{equation}
	where ${j, l}\in \mathcal{N}_i$ belong to the search window $\mathcal{N}_i$ centered at pixel $i$.
	Different from~\cite{rocco17} that consider all possible samples within an image, we constrain search candidates within a local window to reduce matching ambiguity and runtime.
	The similarity score is finally normalized over the search candidates to reliably prune incorrect matches by down-weighting the influence of features that have multiple high scores~\cite{rocco18}.
	Note that $A$ and $B$ represents source ($s$) or target ($t$) images. For instance, ${C}^{ss}$ and ${C}^{tt}$ indicate \emph{self}-similarities computed from the source and target images, respectively.
	${C}^{st}$ is the \emph{cross} similarity between source and target images.
	
	\vspace{-10pt}
	
	\paragraph{Semantic alignment networks}
	Our semantic alignment networks consist of two modules: an alignment module that estimates geometric transformation fields, and an uncertainty module that identifies which regions in an image are likely to be mismatched. 
	
	Taking the cross similarity scores between source and target images as an input, the alignment module based on an encoder-decoder architecture with parameters $\mathbf{W}_A$ estimates locally-varying transformation fields to deal with non-rigid geometric deformations more effectively,
	such that $\tau  = {\cal F}({C}^{st};{{\bf{W}}_A})\in{\mathbb{R}^{H \times W \times 2}}$.
	Different from recent semantic alignment approaches \cite{kim18nips,rocco18nips} that estimate local geometric transformations,
	our alignment module employs the detected landmarks as an additional guidance information to focus more on the salient parts of the objects.

	Additionally, inspired by the probabilistic learning model~\cite{kendall18,kendall2017uncertainties}, we formulate an uncertainty module that predicts how accurately the correspondences will be established at a certain image location.
	The predicted unreliable matches are prevented from being utilized during joint learning process to improve the robustness of our model against possible occlusions or ambiguous matches.
	Unlike existing methods~\cite{novotny17learning,novotny18self,klodt18,kendall18} where the uncertainty map is inferred from an input image,
	our uncertainty module leverages the matching score volume ${C}^{st}$ to provide more informative cues, as in the approaches for confidence estimation in stereo matching~\cite{kim2019unified}.
	Concretely, a series of convolutional layers with parameters $\mathbf{W}_C$ are applied to predict the uncertainty map $\sigma$ from matching similarity scores $C^{st}$ such that $\sigma  = {\cal F}({C}^{st};{{\bf{W}}_C}) \in {\mathbb{R}^{H \times W \times 1}}$.
	\vspace{-10pt}
	
	\paragraph{Landmark detection networks}

	To enable our landmark detection networks to focus on more discriminative regions of the object, we explicitly supply local structures of an image by leveraging self-similarity scores ${C}^{ss}$ and ${C}^{tt}$ computed within a local window, as examplified in~\figref{img:4}.
	This is different from existing methods~\cite{zhang18,jakab18} that employ only convolutional features of images and thus often fail to detect semantically meaningful landmarks under challenging conditions.
	
	Formally, we concatenate the extracted features $F^s$ and $F^t$ with self-similarity scores ${C}^{ss}$ and ${C}^{tt}$ respectively,
	and then pass them through the decoder style networks with parameters $\mathbf{W}_D$ to estimate $K+1$ detection score maps for $K$ landmarks and one background, such that $\phi = \mathcal{F}(F \bigcup {C} ;\mathbf{W}_D) \in \mathbb{R}^{H \times W \times (K+1)}$ where $\bigcup$ denotes a concatenation operator.
	The softmax layer is applied at the end of the networks to transform raw score maps into probability maps by normalizing across the $K+1$ channels,
	\begin{equation}\label{equ:4}
		{\psi ^k_i} = {\exp ({\phi ^k_i})}/{{\sum\nolimits_{m = 0}^{K} {\exp ({\phi ^m_i})} }},
	\end{equation}
	where $\phi^k$ is the score map of the $k^{th}$ landmark.
	The spatial coordinate of the $k^{th}$ landmark is then computed as an expected value over the spatial coordinate $i$ weighted by its probability ${\psi_{i}^k}$, similar to the soft argmax operator in~\cite{kendall2017end}:
	\begin{equation}\label{equ:5}
		{\hat{\psi} ^k} = \sum\nolimits_{i} {i \cdot \psi _i^k} /\sum\nolimits_{i} {\psi _i^k}.
	\end{equation}
	This layer is differentiable, enabling us to formulate loss functions with respect to the landmark coordinates, which will be described in the next section.
	
	\begin{figure}
		\centering
		\renewcommand{\thesubfigure}{}
		\subfigure[(a)]
		{\includegraphics[width=0.245\linewidth]{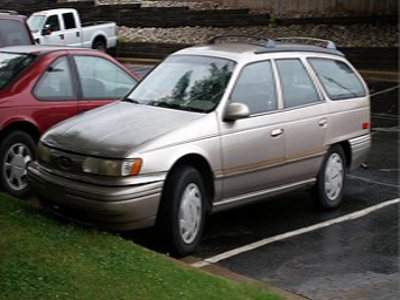}}\hfill
		\subfigure[(b)]
		{\includegraphics[width=0.245\linewidth]{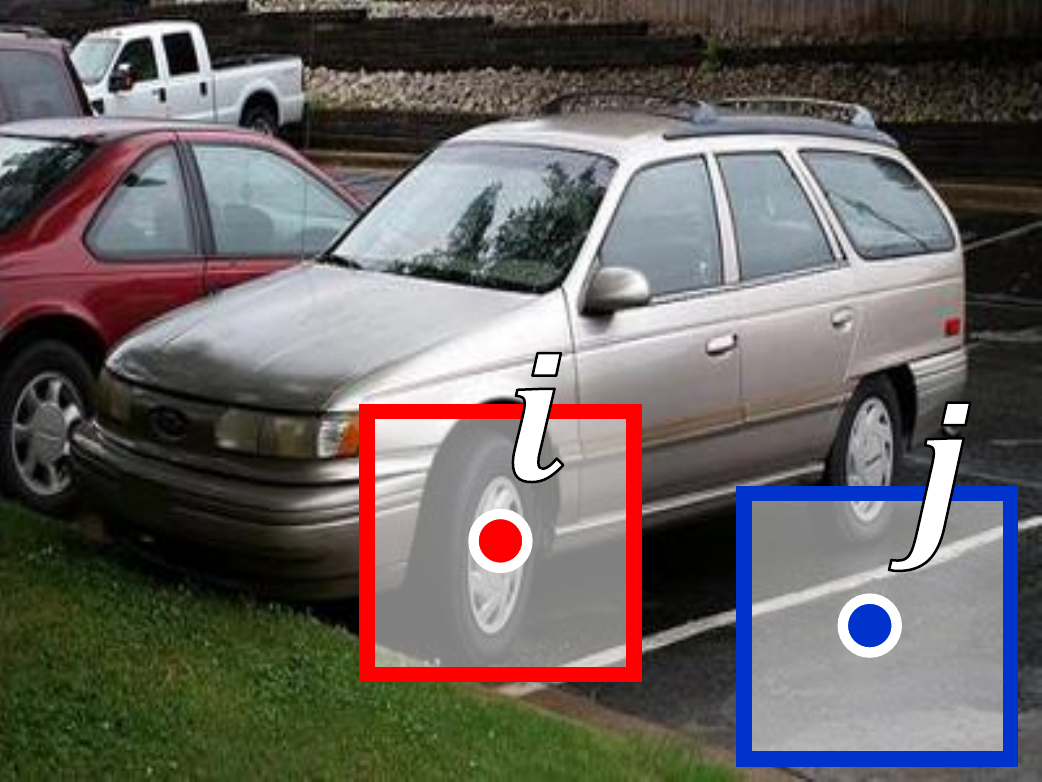}}\hfill
		\subfigure[(c)]
		{\includegraphics[width=0.245\linewidth]{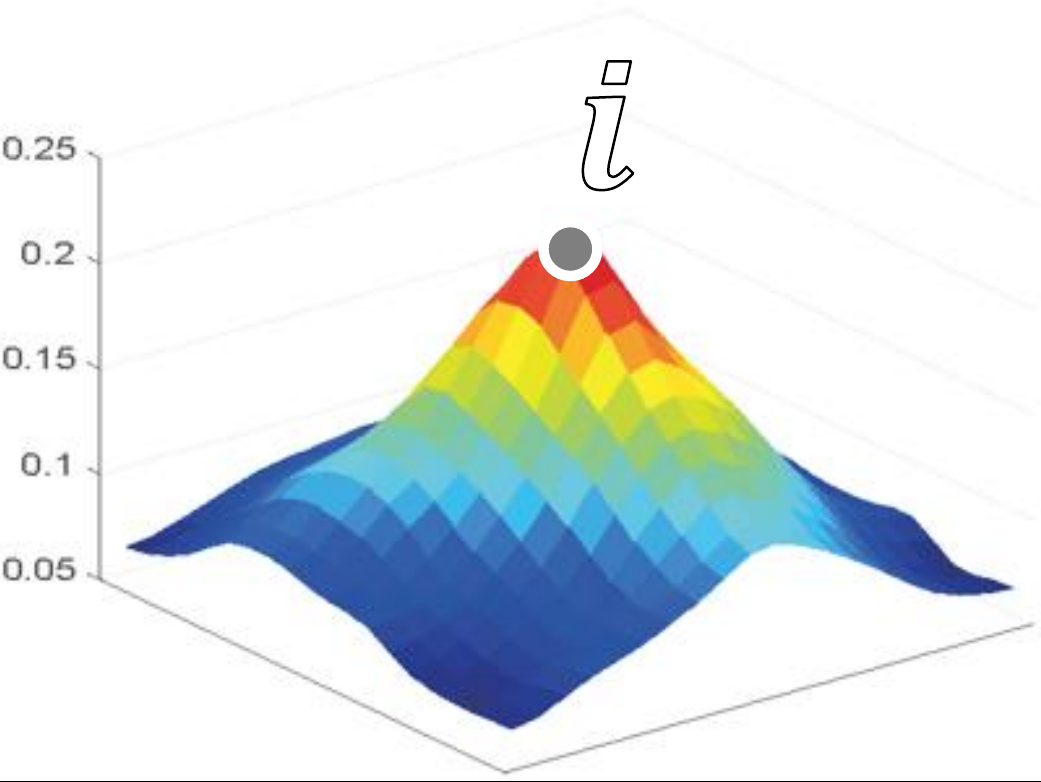}}\hfill
		\subfigure[(d)]
		{\includegraphics[width=0.245\linewidth]{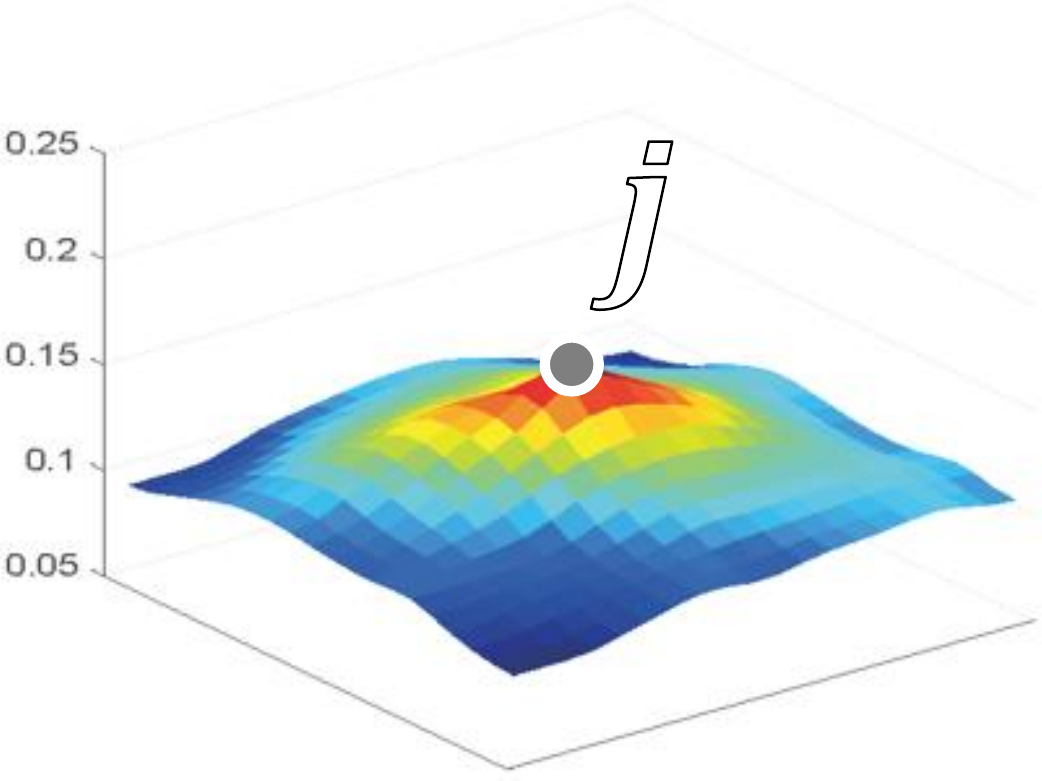}}\hfill
		\vspace{-1pt}
		\caption{Visualization of the effectiveness of self-similarity: (a) an image, (b) arbitrary two coordinates, $i$ and $j$, (c) ${C}^{ss}_i$, and (d) ${C}^{ss}_j$. ${C}^{ss}$ has high variance at more discriminative regions, providing local structure information to landmark detection networks.}\label{img:4}\vspace{-10pt}
	\end{figure}
	
	\subsection{Objective Functions}\label{sec:4.2}
	\paragraph{Loss for semantic alignment networks}
	Our semantic alignment networks are learned using weak image-level supervision in a form of matching image pairs. Concretely, we start with recent weakly-supervised learning techniques proposed in~\cite{kim2018fcss,kim18nips}.
	Under the assumption that corresponding features of source and target images are identical,
	they cast the semantic alignment into a classification task such that the networks can learn the geometric field as a hidden variable over a set of transformation candidates.
	However, this strict assumption is often violated, \eg around occlusions, textureless regions and background clutters, thus requiring additional object location priors to penalize regions where the assumption is invalid.
	
	To address this, we propose to identify \emph{unreliable matches} through the probabilistic formulation of cross-entropy loss such that
	\begin{equation}\label{equ:6}
		{\cal L}{_\mathrm{A}}(\tau,\sigma) = \sum\limits_i (-\sum\limits_{j \in {{\cal M}_i}} \frac{s_j^*}{\sigma_i}\log (s_{i,j}(\tau_i)) + \log {\sigma _i}),
	\end{equation}
	where $\sigma$ is the predicted uncertainty map with parameters $\mathbf{W}_C$ and $s_{i,j}(\tau)$ is a softmax probability defined as
	\begin{equation}\label{equ:7}
		s_{i,j}(\tau) = \frac{{\exp ({<F^s_i,[\tau  \circ {F^t}]_j>})}}
		{{\sum\limits_{l\in \mathcal{M}_i} {\exp ({<F^s_i,[\tau  \circ {F^t}]_l>})} }}.
	\end{equation}
	For $j \in \mathcal{M}_i$, a class label $s^*_j$ is set to $1$ if $j=i$, and $0$ otherwise such that a center point $i$ becomes a positive sample while other points within $\mathcal{M}_i$ are negative samples.
	By dividing the cross entropy loss with the predicted uncertainty map $\sigma$, we can penalize unreliable matches and avoid them to disrupt the loss function. The $\log \sigma$ serves	as a regularization term to prevent $\sigma$ them from becoming too large.
	\vspace{-10pt}
	
	\paragraph{Losses for landmark detection networks}
	Following~\cite{thewlis17iccv,zhang18,suwajanakorn2018discovery}, our landmark detection networks are designed to meet the two common characteristics of landmarks, such that each probability map $\hat{\psi}$ should concentrate on a discriminative local region and, at the same time, distributed at different parts of an object.
	
	The first constraint is used to define a concentration loss ${{\cal L}_\mathrm{con}(\psi)}$ that minimizes the variance over the spatial coordinate $i$ with respect to the landmark coordinate $\phi$~\cite{zhang18}:
	\begin{equation}\label{equ:8}
		{{\cal L}_\mathrm{con}(\psi)} = \sum\limits_k({\sum\limits_i {{{(i -\hat{\psi}^k)}^2} \cdot \psi _i^k} /\sum\limits_i { \psi _i^k}}).
	\end{equation}
	For the second constraint, we define a hinge embedding loss that encourages the landmarks to be far away than a margin $c$~\cite{suwajanakorn2018discovery}, such that
	\begin{equation}\label{equ:9}
		{{\cal L}_\mathrm{sep}(\psi)} = \sum\limits_k {\sum\limits_{k' \ne k} {\max (0,c - ||{\hat{\psi} ^k} - {\hat{\psi} ^{k'}}||^2)}}.
	\end{equation}
	
	\noindent A final loss for the landmark detection networks is defined as a weighted sum of concentration and separation loss, such that
	${{\cal L}_\mathrm{D}(\psi)} = {\lambda _\mathrm{con}}{{\cal L}_\mathrm{con}(\psi)} + {\lambda _\mathrm{sep}}{{\cal L}_\mathrm{sep}(\psi)}$.
	
	Note that similar loss functions are used in the landmark detection literatures~\cite{thewlis17iccv,zhang18}, but our method is different in that more realistic supervisory signals for training the landmark detection networks are provided from the semantic alignment networks.\vspace{-10pt}
	
	\begin{figure}
		\centering
		\renewcommand{\thesubfigure}{}
		\subfigure[(a)]
		{\includegraphics[width=0.245\linewidth]{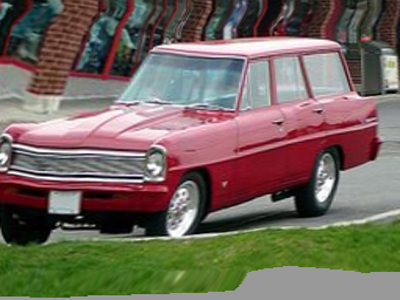}}\hfill
		\subfigure[(b)]
		{\includegraphics[width=0.245\linewidth]{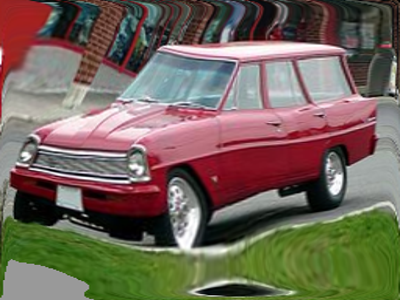}}\hfill
		\subfigure[(c)]
		{\includegraphics[width=0.245\linewidth]{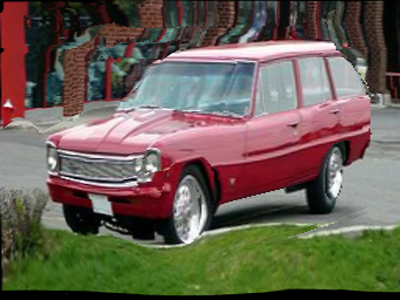}}\hfill
		\subfigure[(d)]
		{\includegraphics[width=0.245\linewidth]{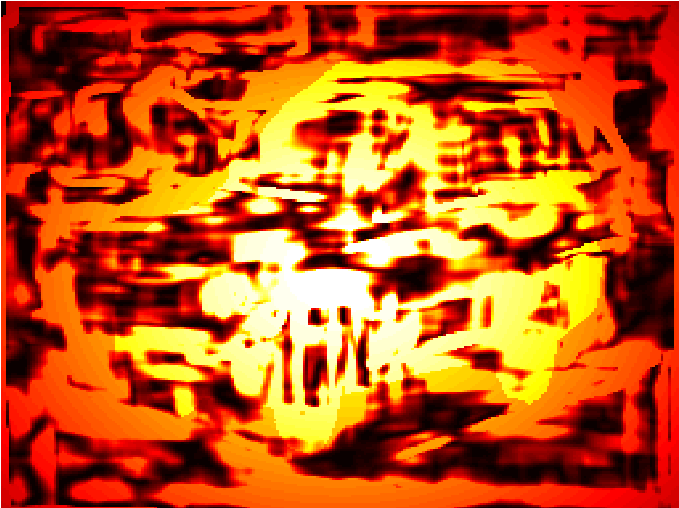}}\hfill
		\vspace{-1pt}
		\caption{Visualization of the effectiveness of probabilistic learning formulation: warped results using correspondences learned (a) from \equref{equ:2} with synthetic supervisions, (b) from \equref{equ:6} without probabilistic formulation, (c) from \equref{equ:6}, and (d) uncertainty map where the darker pixels represent high degree of uncertainty.}\label{img:4}\vspace{-10pt}
	\end{figure}
	
	\paragraph{Loss for joint training}
	Here, we integrate two independent learning processes into a single model by formulating an additional constraint for joint training.
	We apply the outputs of two tasks to a joint distance function as a form of
	\begin{equation}\label{equ:10}
	{L_\mathrm{J}}(\psi ,\tau,\sigma) = \sum\limits_k {\sum\limits_i} \frac{1}{{{\sigma _i}}}||{\psi} _i^k({{\bf{x}}^s}) - \tau \circ {\psi} _i^k({{\bf{x}}^t})||^2.
	\end{equation}
	By imposing the consistency constraint between the landmark detection and semantic alignment, the joint loss function allows us to mutually take advantage of guidance information from both tasks, boosting the performance and addressing the lack of training data in a principled manner.
	Furthermore, we mitigate the adverse impact of unreliable matches in the joint learning process by discounting the contributions of them with the predicted uncertainty map $\sigma_i$.
	Note that instead of landmark coordinates $\hat{\psi}$ in~\equref{equ:10}, the probability map $\psi$ is utilized for a stronger spatial consistency between two tasks.
	A final objective can be defined as a weighted summation of the presented three losses:
	\begin{equation}\label{equ:11}
	{{\mathcal L}_\mathrm{JDA}(\psi,\tau,\sigma)} = \lambda_\mathrm{D}{{\cal L}_\mathrm{D}(\psi)} + \lambda_\mathrm{A}{{\cal L}_\mathrm{A}(\tau,\sigma)}+\lambda_\mathrm{J}{{\cal L}_\mathrm{J}(\psi,\tau,\sigma)}.
	\end{equation}
	\vspace{-15pt}
	
		\begin{figure}[t]
			\centering
			\renewcommand{\thesubfigure}{}
			\subfigure[]
			{\includegraphics[width=0.245\linewidth]{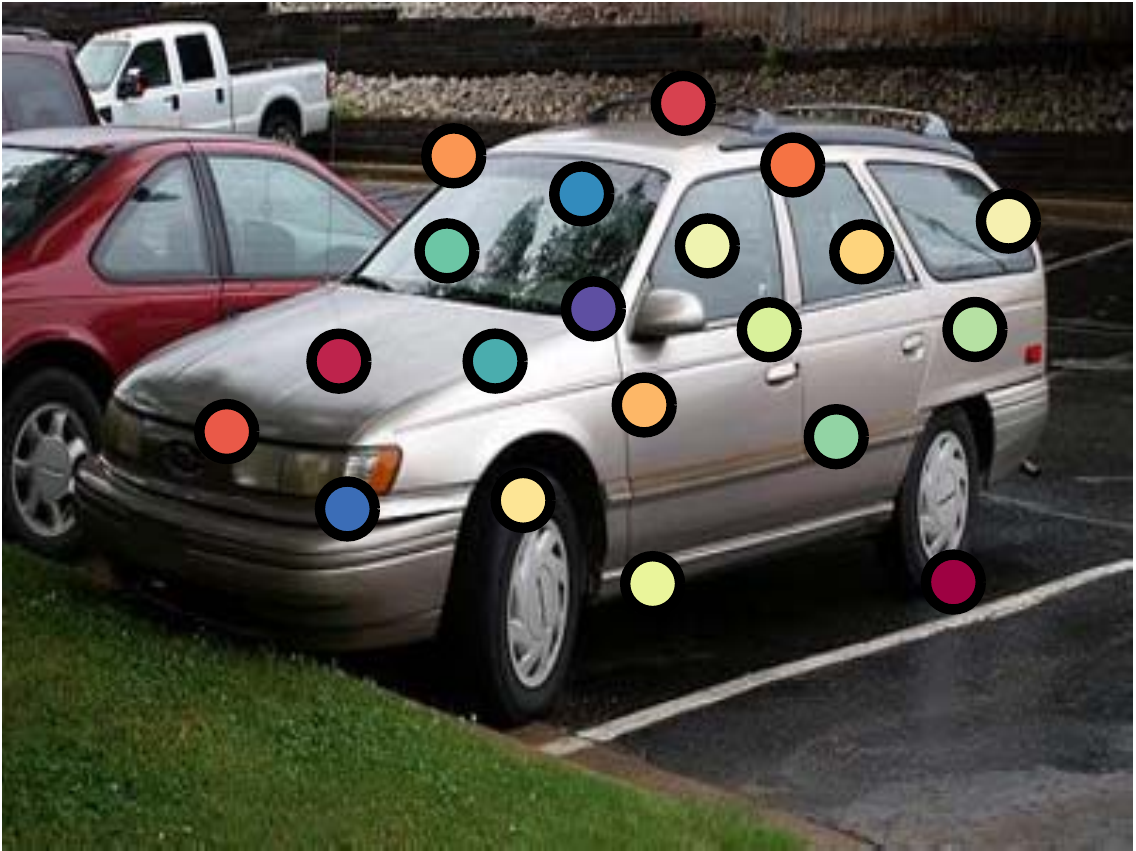}}\hfill
			\subfigure[]
			{\includegraphics[width=0.245\linewidth]{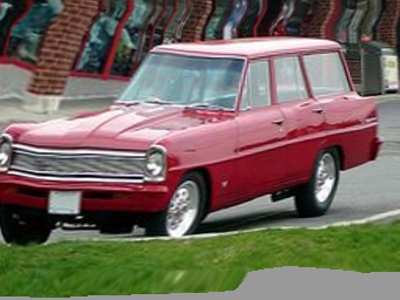}}\hfill
			\subfigure[]
			{\includegraphics[width=0.245\linewidth]{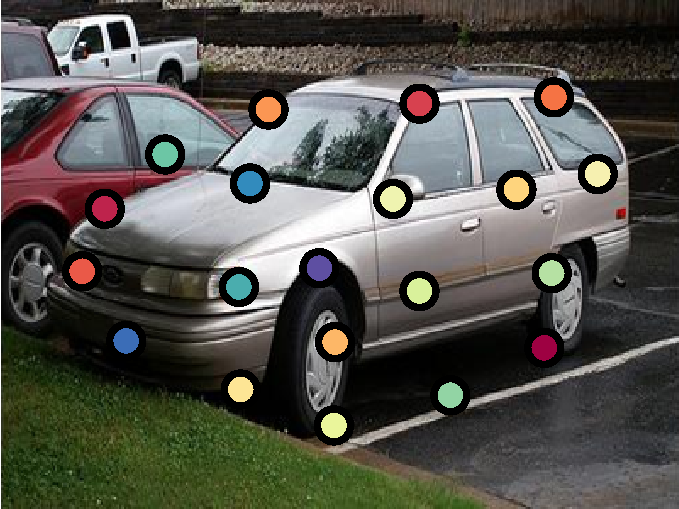}}\hfill
			\subfigure[]
			{\includegraphics[width=0.245\linewidth]{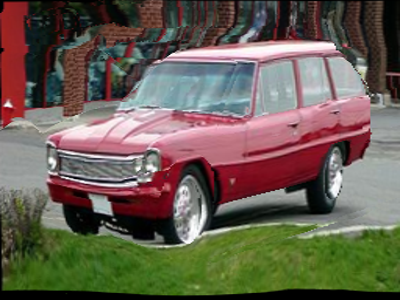}}\hfill
			\vspace{-21.5pt}
			\subfigure[(a)]
			{\includegraphics[width=0.245\linewidth]{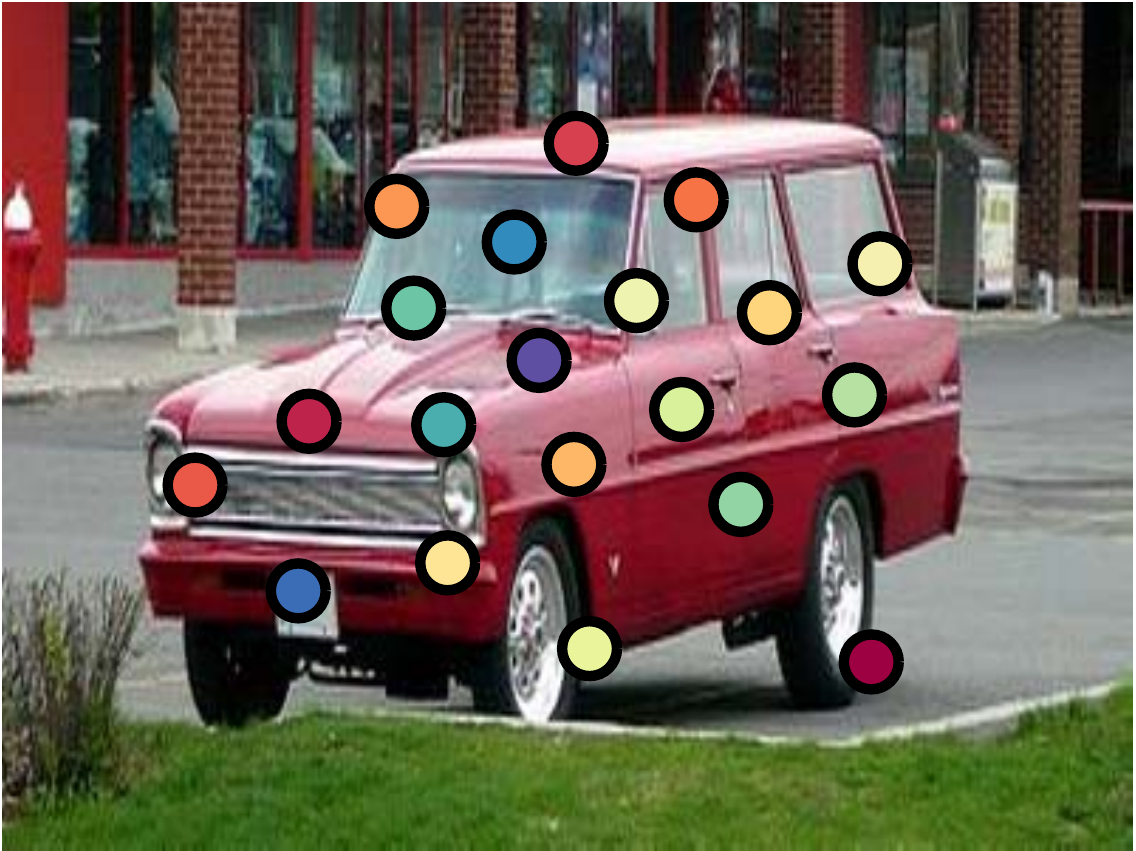}}\hfill
			\subfigure[(b)]
			{\includegraphics[width=0.245\linewidth]{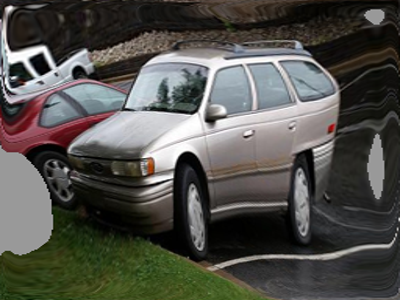}}\hfill
			\subfigure[(c)]
			{\includegraphics[width=0.245\linewidth]{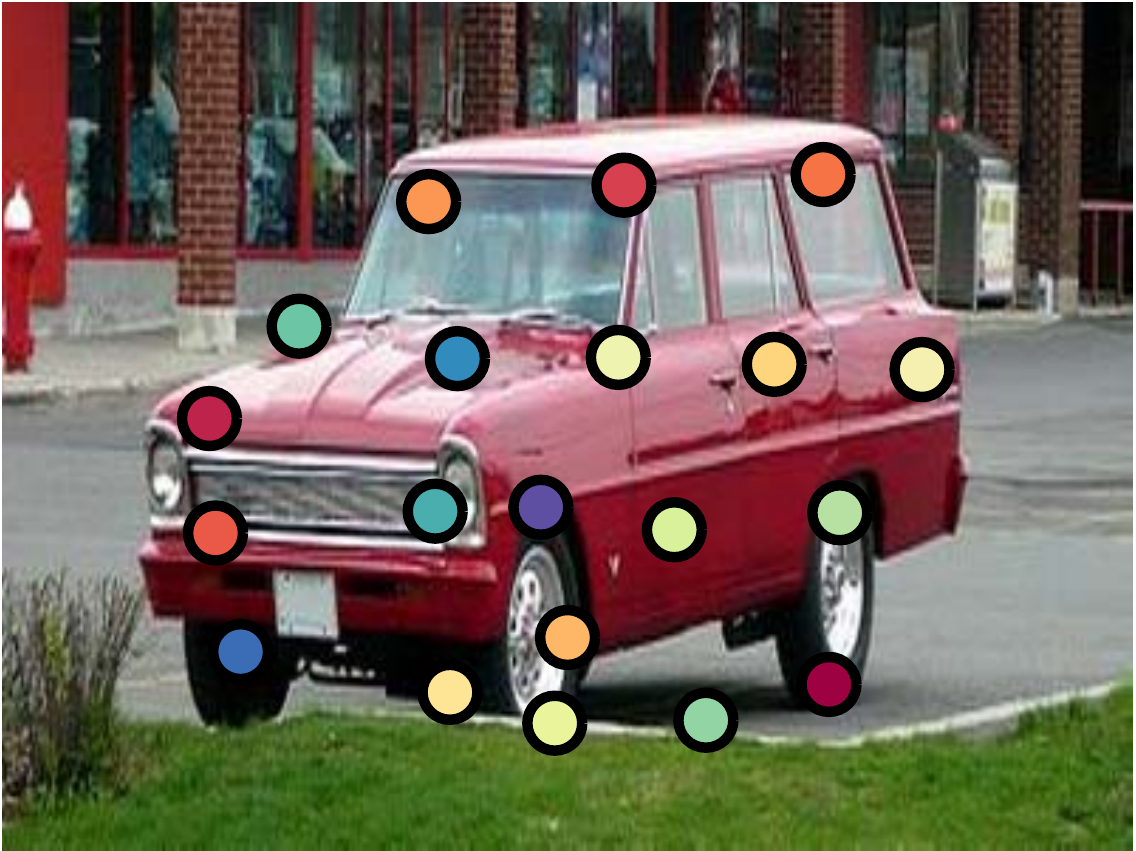}}\hfill
			\subfigure[(d)]
			{\includegraphics[width=0.245\linewidth]{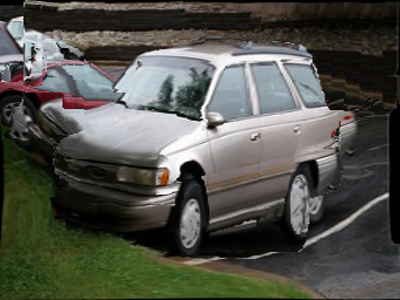}}\hfill
			\vspace{-3pt}
			\caption{The effectiveness of the proposed joint learning framework: detected landmarks and aligned images (a), (b) when learned separately, and (c), (d) when learned jointly .}\label{img:6}\vspace{-10pt}
		\end{figure}
		
	\subsection{{Training}}
	\paragraph{Alternative Optimization}
	To optimize the landmark detection and semantic alignment networks in a mutually reinforced way, we learn the landmark detection networks and semantic alignment networks in an alternating fashion.
	For better initialization, we first pretrain both networks independently with synthetically generated image pairs, similar to~\cite{rocco18}.
	Randomly perturbed images are generated by applying global affine or TPS transformation to the original images from the Pascal VOC 2012 segmentation dataset~\cite{everingham2010pascal}, and utilize these image pairs for learning each networks with loss functions~\equref{equ:1} and~\equref{equ:2}.
	In sequence, we finetune both pretrained networks in an end-to-end manner for semantically similar images pairs from the JLAD dataset described in the following section. Specifically, the network parameters $\{\mathbf{W}_F,\mathbf{W}_A,\mathbf{W}_C\}$ are optimized for semantic alignment by setting $\{\lambda_D,\lambda_A,\lambda_J\}$ as $\{1,10,10\}$, and $\{\mathbf{W}_F,\mathbf{W}_D\}$ for landmark detection by setting $\{\lambda_D,\lambda_A,\lambda_J\}$ as $\{10,1,100\}$.
	We iterate this process until the final objective converges.
	\vspace{-10pt}
	
	\paragraph{{JLAD Dataset}}
	To learn our networks with the proposed consistency constraint~\equref{equ:11}, large-scale semantically similar image pairs are required, but existing public datasets are limited quantitatively.
	To overcome this, we introduce a new dataset that contains a larger number of challenging image pairs, called JLAD dataset. The images and keypoint annotations are sampled and refined from the original ones of PASCAL 3D benchmark~\cite{xiang2014beyond} and MAFL dataset~\cite{zhang2014facial}.
	For each object category in PASCAL 3D dataset~\cite{xiang2014beyond} which provides about 36,000 images for 12 categories, we first	preprocessed their images to contain only a single object.
	Specifically, the images are cropped according to the provided object bounding box annotations, including margins for background clutters. Then using the ground-truth viewpoint annotations such as azimuth and elevation angles, we sampled about 1,000 image pairs for each category.
	For human
	faces, we sampled image pairs randomly from MAFL
	dataset~\cite{zhang2014facial} excluding testing set without considering geometric
	constraints since their images are already cropped
	and aligned.
	We used the split which divides the collected
	image pairs into roughly 70 percent for training, 20 percent
	for validation, and 10 percent for testing.
	
	\section{Experimental Results}
	\subsection{Experimental Settings}
	For feature extraction, we used the ImageNet-pretrained ResNet~\cite{He16}, where the activations are sampled after pooling layers such as `conv4-23' for ResNet-101~\cite{He16}.
	Margin $c$ is set to be 0.05, 0.03, 0.02 for detecting 10, 15, 30 landmarks respectively.
	The radius of the search space for $\mathcal{N}_i$ is set to 5, which is equivalent to $40 \times 40$ window at the original resolution.
	Following~\cite{kendall18}, our uncertainty networks are formulated to predict log variance of uncertainty,~\ie $\mathrm{log}\sigma$, to avoid a potential division of~\equref{equ:6} by zero.  
	During alternative optimization, we set the maximum number of alternation to 4 to avoid overfitting.
	We used ADAM optimizer~\cite{kingma2014adam} with $\beta_1=0.9$ and $\beta_2=0.999$. We set the training
	batch size to 16. A learning rate initially set to $10^{-3}$ and
	decreases to $10^{-4}$ and $10^{-5}$ later.
		\begin{table}[!t]
			\begin{center}
				\begin{tabular}{ >{\raggedright}m{0.30\linewidth}
						>{\centering}m{0.3\linewidth} >{\centering}m{0.3\linewidth}}
					\hlinewd{0.8pt}
					\multicolumn{1}{l|}{\multirow{2}{*}{Methods}} &\multicolumn{1}{ c|}{Alignment acc.} &Detection acc.\tabularnewline
					\cline{2-3}
					\multicolumn{1}{l|}{} &\multicolumn{1}{ c|}{PCK@$\alpha=0.1$} &IOD \tabularnewline
					\hline
					\hline
					\multicolumn{1}{l|}{Separate learning}  &\multicolumn{1}{ c|}{63.2} &7.97  \tabularnewline
					\multicolumn{1}{l|}{Iteration 1} &\multicolumn{1}{ c|}{67.0} &7.36 \tabularnewline
					\multicolumn{1}{l|}{Iteration 2} &\multicolumn{1}{ c|}{70.2} &7.16 \tabularnewline
					\multicolumn{1}{l|}{Iteration 3} &\multicolumn{1}{ c|}{72.1} &7.05 \tabularnewline
					\hline
					\multicolumn{1}{l|}{Ours} &\multicolumn{1}{ c|}{\bf{72.7}} &\bf{6.92} \tabularnewline
					\hlinewd{0.8pt}
				\end{tabular}
			\end{center}
			\vspace{-5pt}
			\caption{Ablation study for the effectiveness of the proposed joint learning framework on the JLAD dataset. The accuracies for semantic alignment and object landmark detection are reported with PCK and IOD metrics, respectively.}\label{tab:1}\vspace{-10pt}
		\end{table}
	
	\begin{table*}[t]
		\centering
		\begin{tabular}{ >{\raggedright}m{0.30\linewidth} >{\centering}m{0.04\linewidth} >{\centering}m{0.04\linewidth} >{\centering}m{0.04\linewidth} >{\centering}m{0.04\linewidth}
				>{\centering}m{0.04\linewidth} >{\centering}m{0.04\linewidth} >{\centering}m{0.04\linewidth} >{\centering}m{0.04\linewidth} >{\centering}m{0.04\linewidth} >{\centering}m{0.04\linewidth}
				>{\centering}m{0.04\linewidth} >{\centering}m{0.04\linewidth} >{\centering}m{0.04\linewidth}}
			\hlinewd{0.8pt}
			\multicolumn{1}{l|}{Methods} &aero. &bicy. &boat &bott. &bus &car &chair &d.table &motor. &sofa &train &tv. &\multicolumn{1}{|c}{All} \tabularnewline
			\hline
			\hline
			\multicolumn{1}{l|}{CNNgeo~\cite{rocco17}}  &71.3 &74.4 &44.4 &60.9 &79.6 &83.8 &63.9 &36.6 &72.1 &43.8 &42.5 &48.0 &\multicolumn{1}{|c}{60.1} \tabularnewline
			\multicolumn{1}{l|}{CNNinlier~\cite{rocco18}}  &79.6 &82.9 &54.4 &68.7 &89.5 &88.5 &70.7 &39.2 &79.4 &48.2 &49.4 &51.1 &\multicolumn{1}{|c}{66.8} \tabularnewline
			\multicolumn{1}{l|}{A2Net~\cite{paul18}}  &80.9 &81.4 &53.6 &69.5 &88.6 &89.5 &71.3 &41.2 &78.1 &51.8 &52.0 &51.7 &\multicolumn{1}{|c}{67.5} \tabularnewline
			\multicolumn{1}{l|}{RTNs~\cite{kim18nips}}  &81.5 &85.4 &56.3 &70.8 &87.4 &92.7 &72.3 &43.6 &84.3 &59.8 &55.2 &53.5 &\multicolumn{1}{|c}{70.2} \tabularnewline
			\multicolumn{1}{l|}{NCNet~\cite{rocco18nips}}  &82.4 &85.2 &57.9 &71.2 &88.8 &93.1 &\bf{75.8} &\bf{46.9} &87.8 &57.7 &57.1 &\bf{56.5} &\multicolumn{1}{|c}{71.7} \tabularnewline
			\hline
			\multicolumn{1}{l|}{Ours} &\bf{84.7} &\bf{89.1} &\bf{62.5}  &\bf{74.5} &\bf{90.3} &\bf{93.3} &73.3 &46.7 &\bf{89.4} &\bf{60.7} &\bf{62.1} &56.3 &\multicolumn{1}{|c}{\bf{73.6}} \tabularnewline
			\hlinewd{0.8pt}
		\end{tabular}\vspace{+3pt}
		\caption{Matching accuracy compared to the state-of-the-art semantic alignment techniques over various object categories on the JLAD dataset. The distance threshold of PCK $\alpha$ is set to 0.01.}\label{tab:2}\vspace{-10pt}
	\end{table*}

	In the following, we comprehensively evaluated our framework in comparison to state-of-the-art methods for landmark detection,
	including FPE~\cite{thewlis17iccv}, DEIL~\cite{thewlis17nips}, StrucRep~\cite{zhang18}, CIG~\cite{jakab18}, and for semantic alignment, including CNNgeo~\cite{rocco17}, CNNinlier~\cite{rocco18}, A2Net~\cite{paul18} and NCNet~\cite{rocco18nips}.
	Performance was measured on JLAD dataset and PF-PASCAL~\cite{Ham17} for 12 object categories, and on MAFL dataset~\cite{zhang2014facial} and AFLW dataset~\cite{koestinger2011annotated} for human faces.
	See the supplemental material for more details on the implementation of our system and more qualitative results.
	
	\subsection{Ablation Study}\label{sec:5.3}
	We first analyze the effectiveness of the components within our method.
	The performances of landmark detection and semantic alignment are examined for different numbers of alternative iterations.
    The qualitative and quantitatve assessments are conducted on the testing image pairs of JLAD dataset.
    As shown in~\tabref{tab:1} and~\figref{img:6}, the results of our joint learning model show significant improvements in comparision to separate estimation models that rely on synthetic transformations.
	We also conducted an ablation study by removing the uncertainty prediction model within semantic alignment networks (Ours wo/UM) and the correlation layer within landmark detection network that computes local self-similarties (Ours wo/SS).
	Degraded performance of ``Ours wo/SS'' and ``Ours wo/UM'' in~\tabref{tab:1} and~\tabref{tab:2} highlights the importance of encoding local structure through self-similarities for landmark detection and considering possible ambiguous matches for semantic alignment.
	
	\begin{table}[!t]
		\begin{center}
			\begin{tabular}{ >{\raggedright}m{0.31\linewidth}
					>{\centering}m{0.16\linewidth} >{\centering}m{0.16\linewidth}
					>{\centering}m{0.16\linewidth}}
				\hlinewd{0.8pt}
				\multicolumn{1}{l|}{\multirow{2}{*}{Methods}}
				&\multicolumn{3}{ c }{PCK} \tabularnewline
				\cline{2-4}
				\multicolumn{1}{l|}{} &$\alpha=0.05$ &$\alpha=0.1$ &$\alpha=0.15$\tabularnewline
				\hline
				\hline
				\multicolumn{1}{l|}{CNNgeo~\cite{rocco17}} &36.9 &62.3 &71.4 \tabularnewline
				\multicolumn{1}{l|}{CNNinlier~\cite{rocco18}} &44.1 &68.2 &74.8 \tabularnewline
				\multicolumn{1}{l|}{A2Net~\cite{paul18}} &43.1 &68.4 &74.1 \tabularnewline
				\multicolumn{1}{l|}{RTNs~\cite{kim18nips}} &49.2 &69.3 &76.2 \tabularnewline
				\multicolumn{1}{l|}{NCNet~\cite{rocco18nips}} &50.7 &70.9 &78.1 \tabularnewline
				\hline
				\multicolumn{1}{l|}{Ours wo/UM} &{49.4} &{68.2} &{76.9} \tabularnewline
				\multicolumn{1}{l|}{Ours} &\bf{52.8} &\bf{72.7} &\bf{79.2} \tabularnewline
				\hlinewd{0.8pt}
			\end{tabular}
		\end{center}
		\vspace{-5pt}
		\caption{Matching accuracy compared to the state-of-the-art correspondence techniques on the PF-PASCAL benchmark~\cite{Ham17}.}\label{tab:3}\vspace{-10pt}
	\end{table}
	
	\subsection{Results}\label{sec:5.4}

	\begin{figure*}[t]
		\centering
		\renewcommand{\thesubfigure}{}
		\subfigure[]
		{\includegraphics[width=0.122\linewidth]{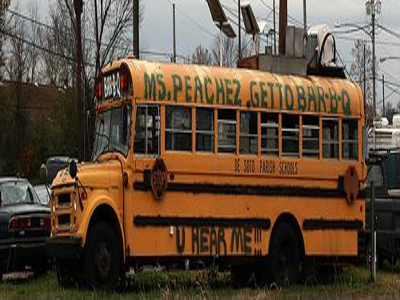}}\hfill
		\subfigure[]
		{\includegraphics[width=0.122\linewidth]{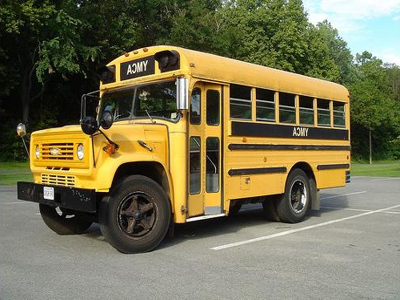}}\hfill
		\subfigure[]
		{\includegraphics[width=0.122\linewidth]{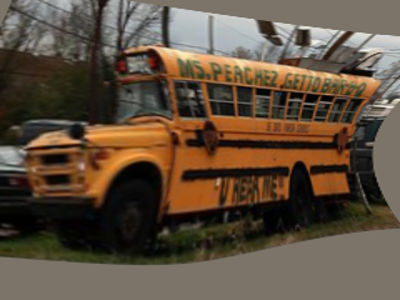}}\hfill
		\subfigure[]
		{\includegraphics[width=0.122\linewidth]{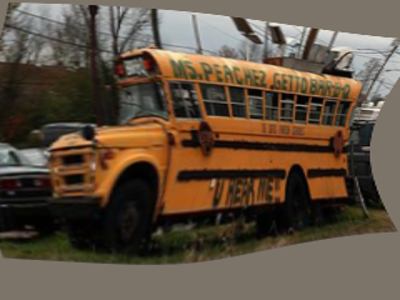}}\hfill
		\subfigure[]
		{\includegraphics[width=0.122\linewidth]{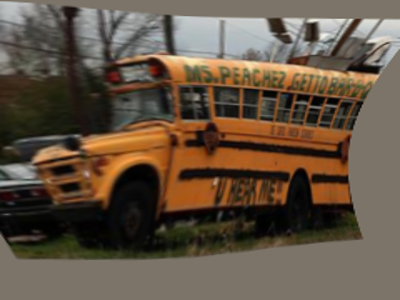}}\hfill
		\subfigure[]
		{\includegraphics[width=0.122\linewidth]{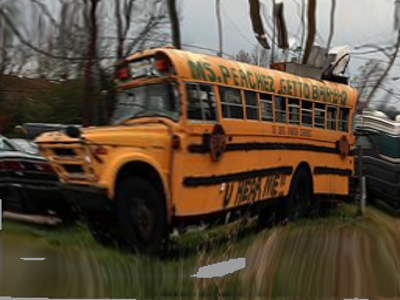}}\hfill
		\subfigure[]
		{\includegraphics[width=0.122\linewidth]{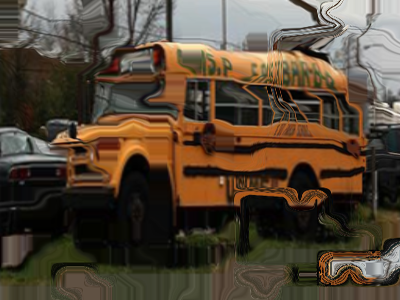}}\hfill
		\subfigure[]
		{\includegraphics[width=0.122\linewidth]{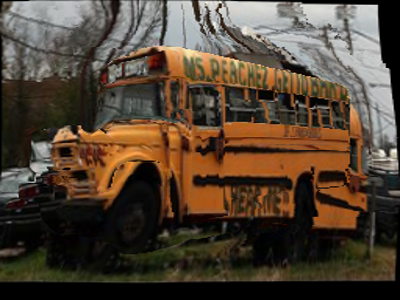}}\hfill
		\vspace{-21.5pt}
		\subfigure[]
		{\includegraphics[width=0.122\linewidth]{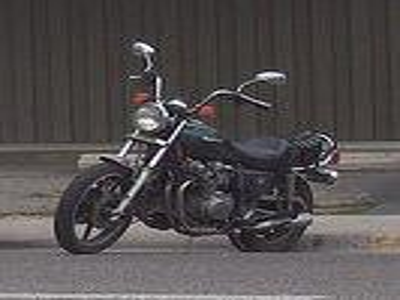}}\hfill
		\subfigure[]
		{\includegraphics[width=0.122\linewidth]{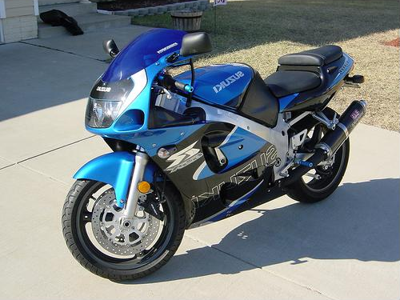}}\hfill
		\subfigure[]
		{\includegraphics[width=0.122\linewidth]{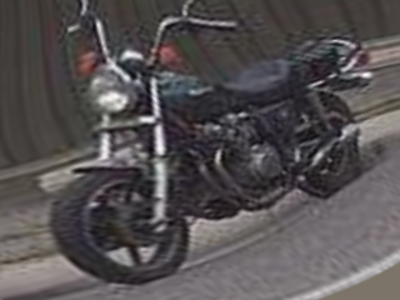}}\hfill
		\subfigure[]
		{\includegraphics[width=0.122\linewidth]{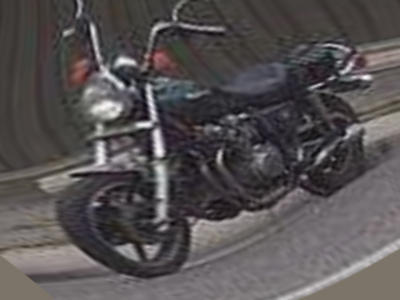}}\hfill
		\subfigure[]
		{\includegraphics[width=0.122\linewidth]{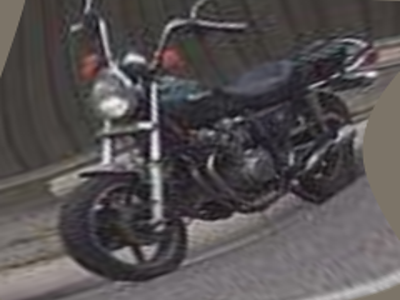}}\hfill
		\subfigure[]
		{\includegraphics[width=0.122\linewidth]{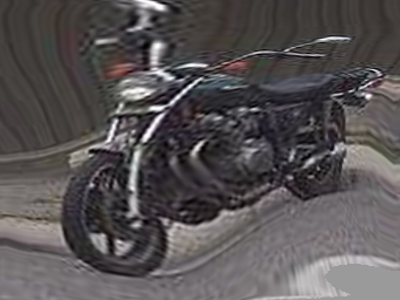}}\hfill
		\subfigure[]
		{\includegraphics[width=0.122\linewidth]{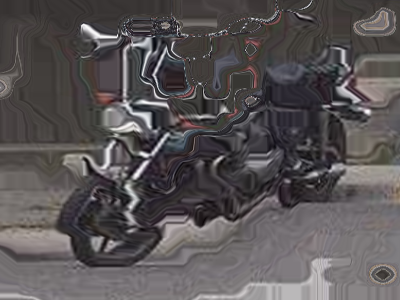}}\hfill
		\subfigure[]
		{\includegraphics[width=0.122\linewidth]{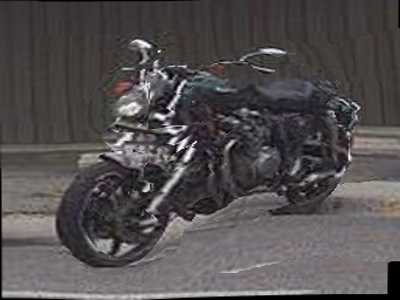}}\hfill
		\vspace{-21.5pt}
		\subfigure[(a)]
		{\includegraphics[width=0.122\linewidth]{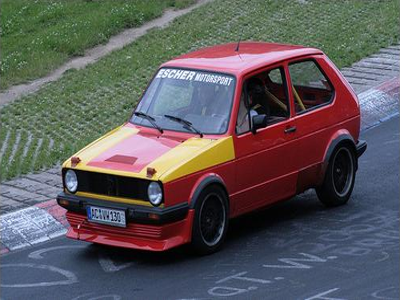}}\hfill
		\subfigure[(b)]
		{\includegraphics[width=0.122\linewidth]{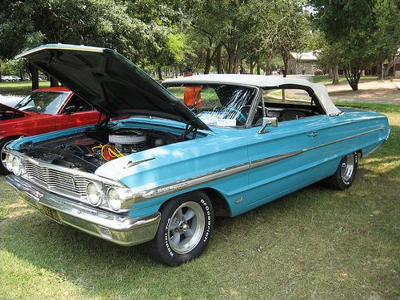}}\hfill
		\subfigure[(c)]
		{\includegraphics[width=0.122\linewidth]{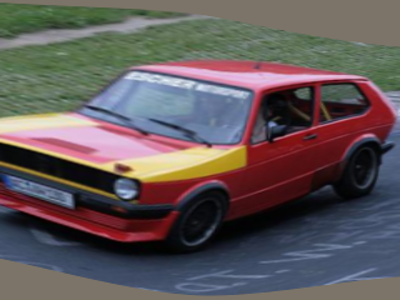}}\hfill
		\subfigure[(d)]
		{\includegraphics[width=0.122\linewidth]{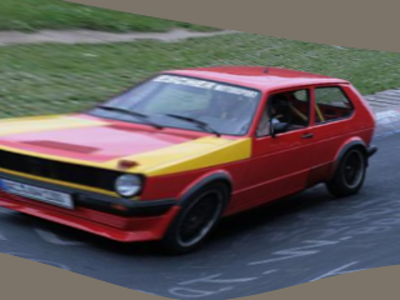}}\hfill
		\subfigure[(e)]
		{\includegraphics[width=0.122\linewidth]{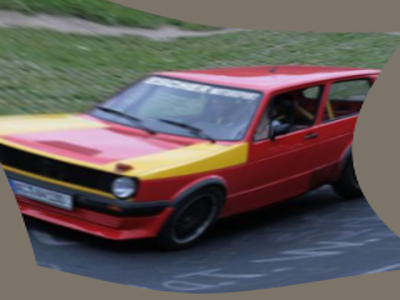}}\hfill
		\subfigure[(f)]
		{\includegraphics[width=0.122\linewidth]{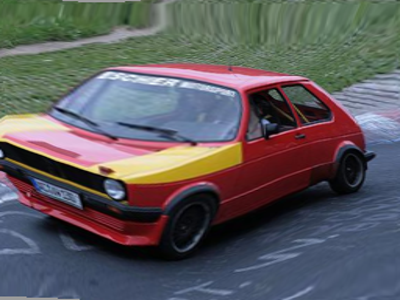}}\hfill
		\subfigure[(g)]
		{\includegraphics[width=0.122\linewidth]{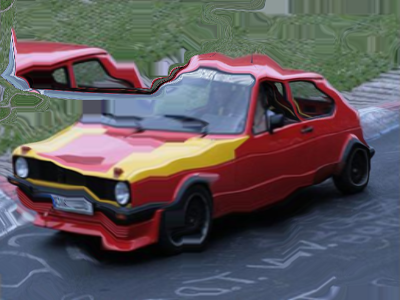}}\hfill
		\subfigure[(h)]
		{\includegraphics[width=0.122\linewidth]{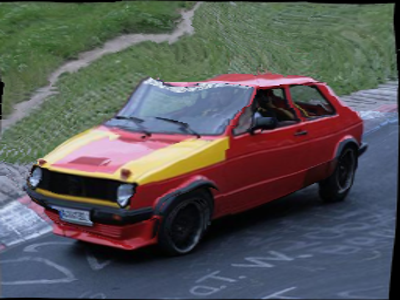}}\hfill
		\vspace{-3pt}
		\caption{Qualitative results of the semantic alignment on the JLAD dataset: (a) source image, (b) target image,
			(c) CNNgeo \cite{rocco17}, (d) CNNinlier \cite{rocco18}, (e) A2Net \cite{paul18}, (f) RTNs~\cite{kim18nips}, (g) NCNet \cite{rocco18nips}, and (h) Ours.
			The source images were warped to the target images using correspondences.}\label{img:7}\vspace{-5pt}
	\end{figure*}

	\begin{figure*}[t]
		\centering
		\renewcommand{\thesubfigure}{}
		\subfigure[]
		{\includegraphics[width=0.122\linewidth]{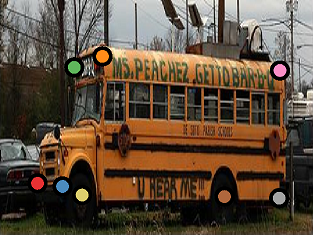}}\hfill
		\subfigure[]
		{\includegraphics[width=0.122\linewidth]{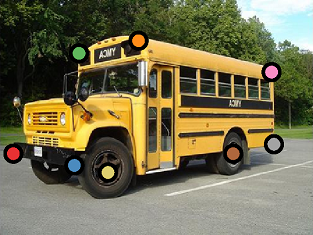}}\hfill
		\subfigure[]
		{\includegraphics[width=0.122\linewidth]{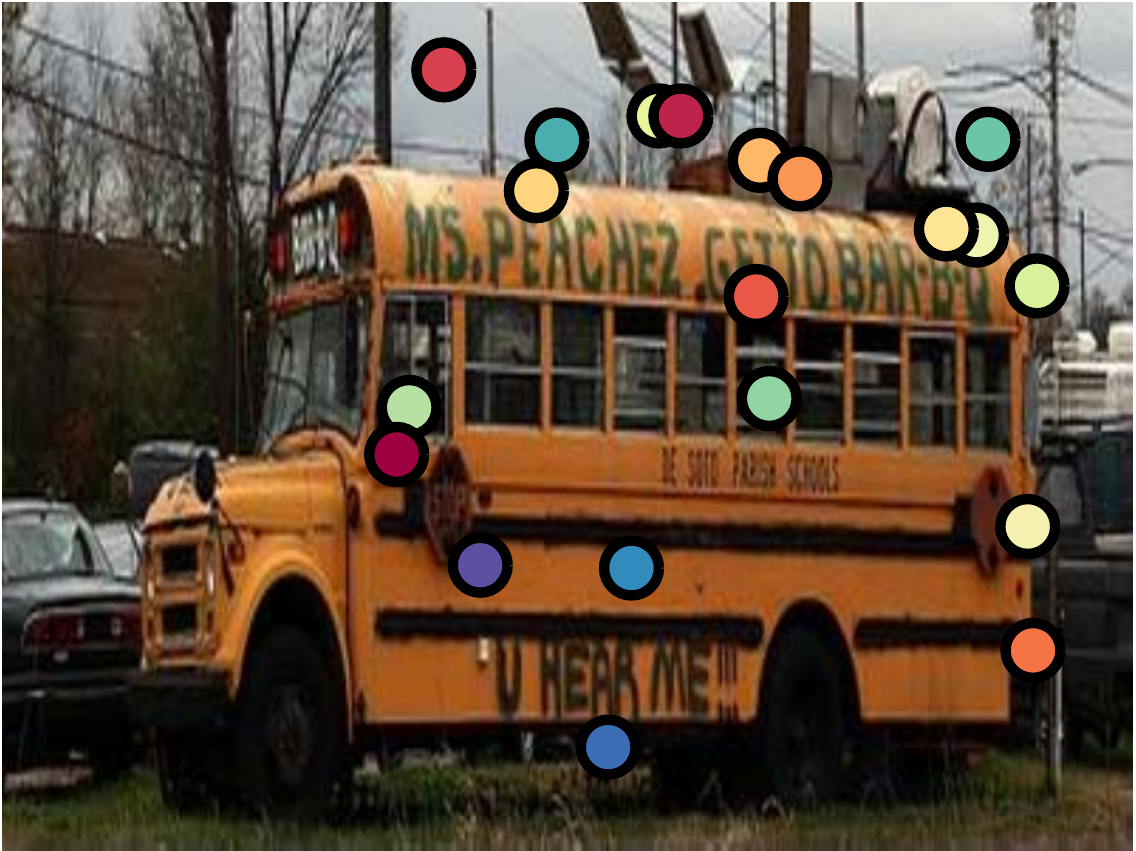}}\hfill
		\subfigure[]
		{\includegraphics[width=0.122\linewidth]{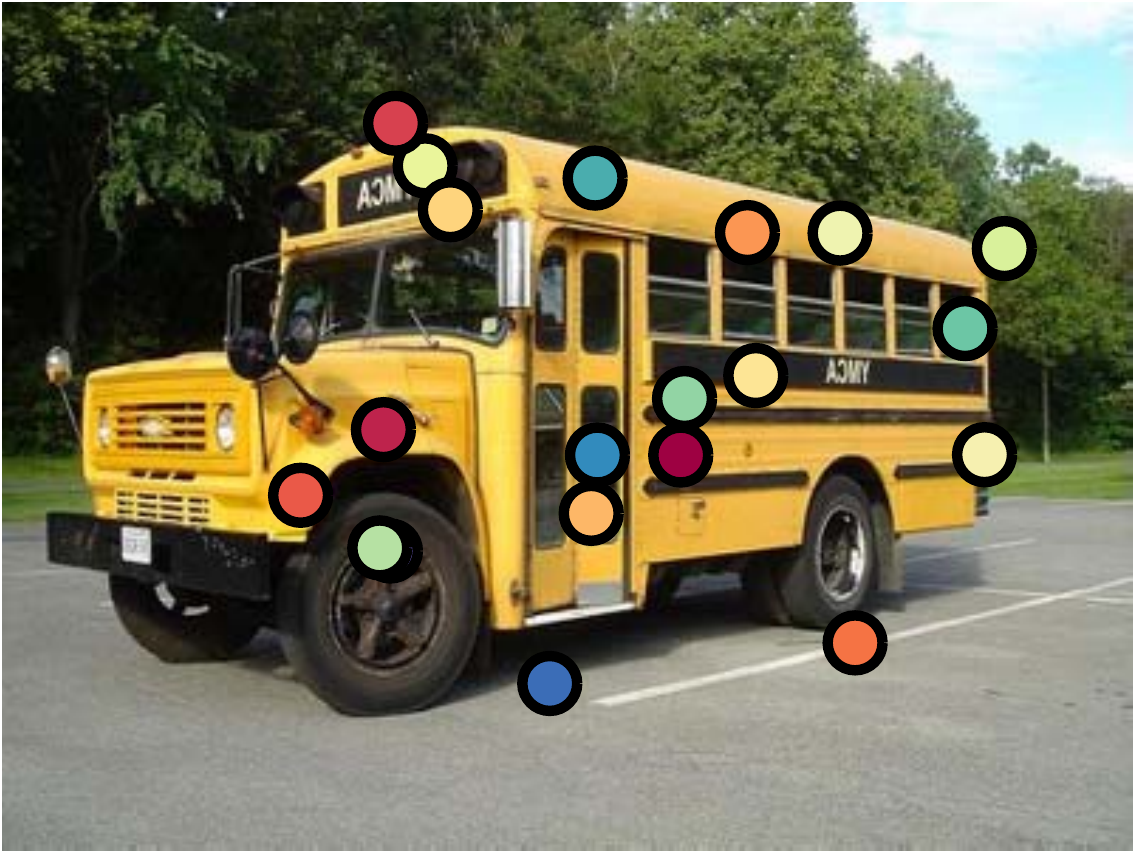}}\hfill
		\subfigure[]
		{\includegraphics[width=0.122\linewidth]{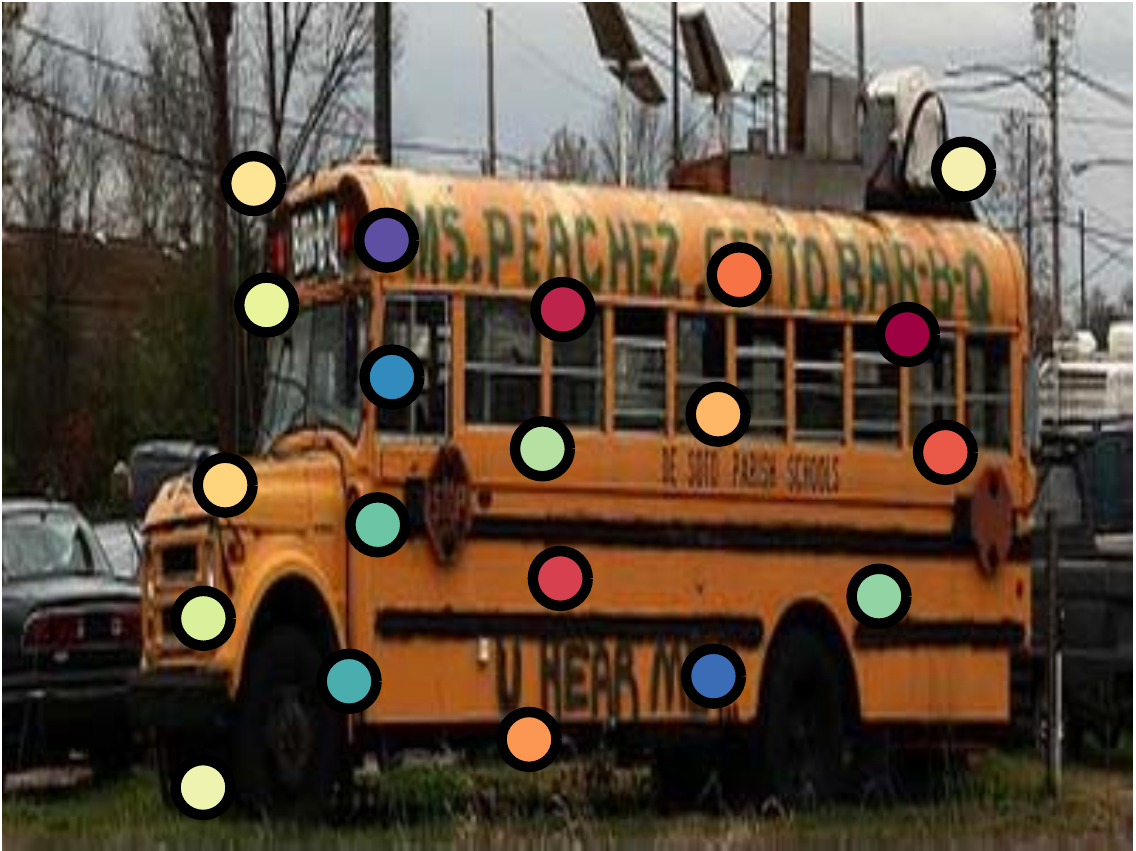}}\hfill
		\subfigure[]
		{\includegraphics[width=0.122\linewidth]{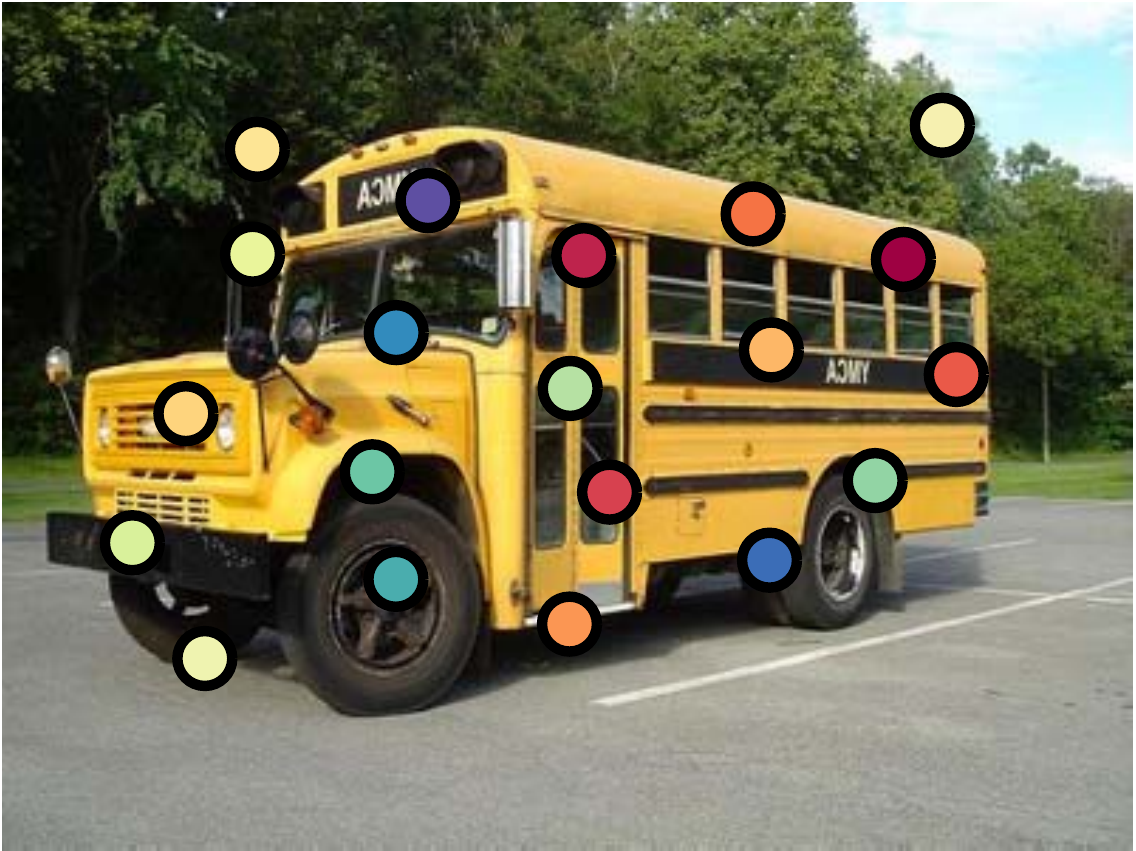}}\hfill
		\subfigure[]
		{\includegraphics[width=0.122\linewidth]{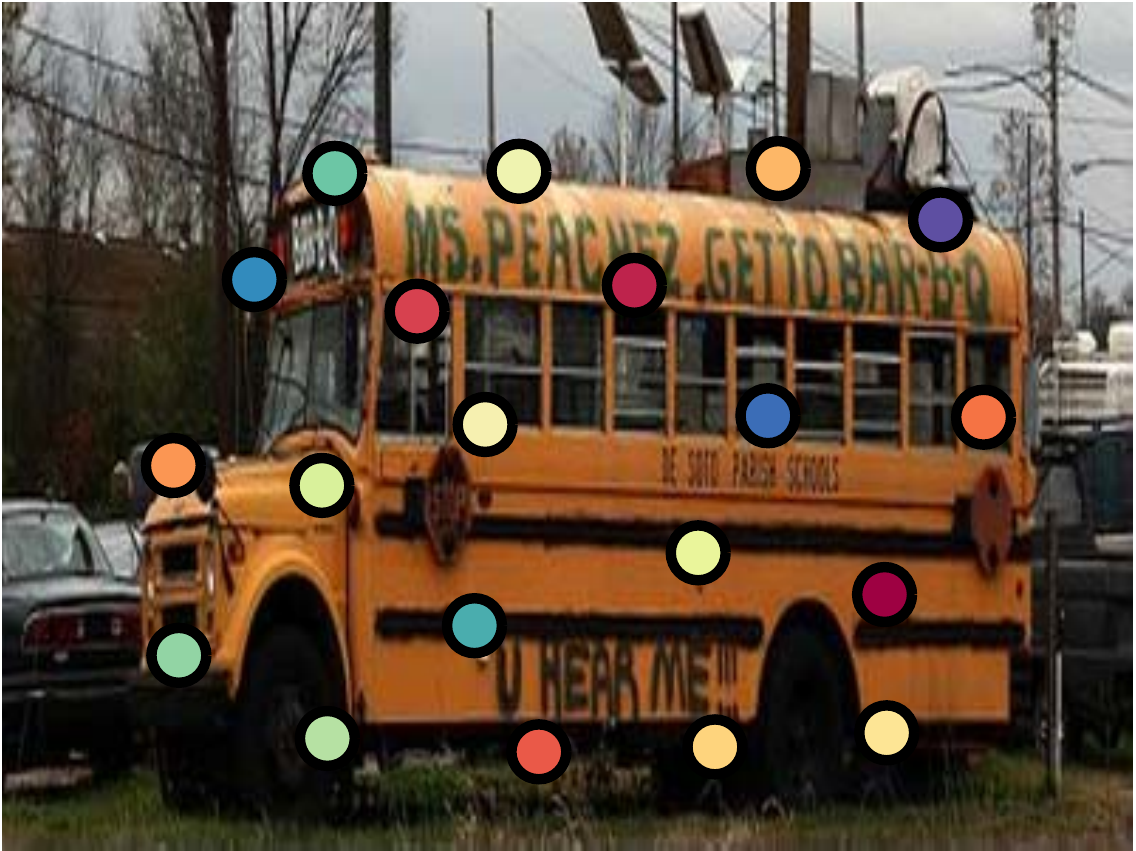}}\hfill
		\subfigure[]
		{\includegraphics[width=0.122\linewidth]{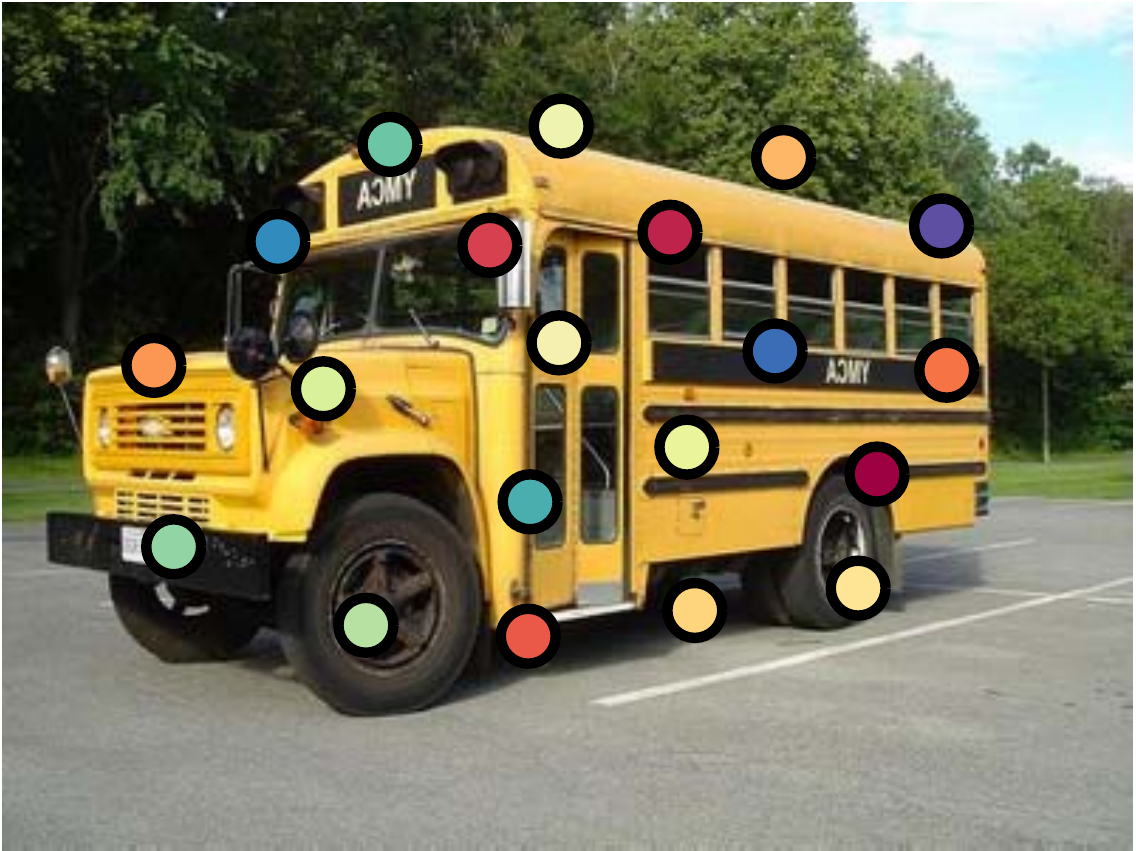}}\hfill
		\vspace{-21.5pt}
		\subfigure[]
		{\includegraphics[width=0.122\linewidth]{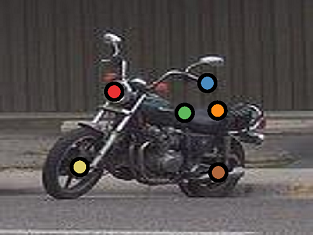}}\hfill
		\subfigure[]
		{\includegraphics[width=0.122\linewidth]{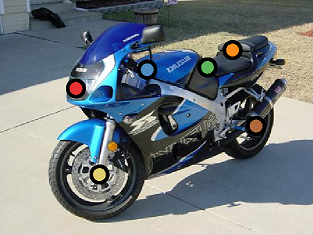}}\hfill
		\subfigure[]
		{\includegraphics[width=0.122\linewidth]{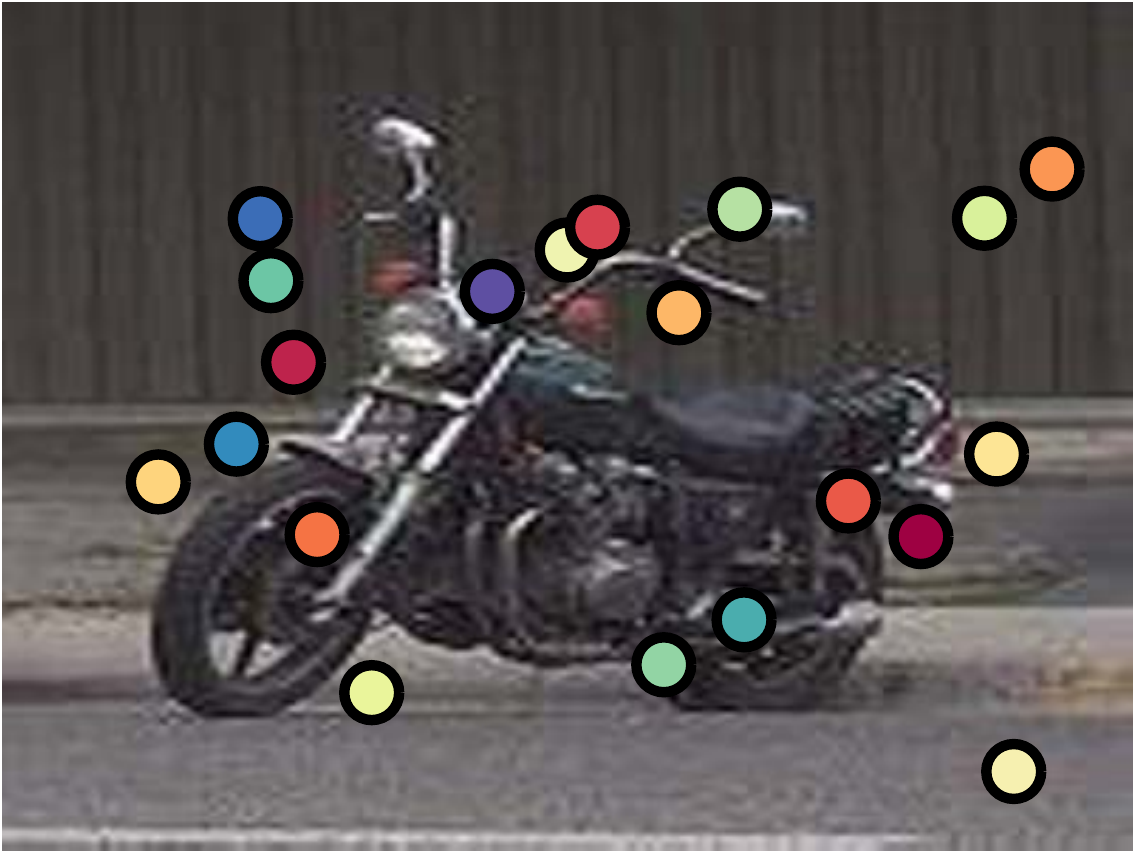}}\hfill
		\subfigure[]
		{\includegraphics[width=0.122\linewidth]{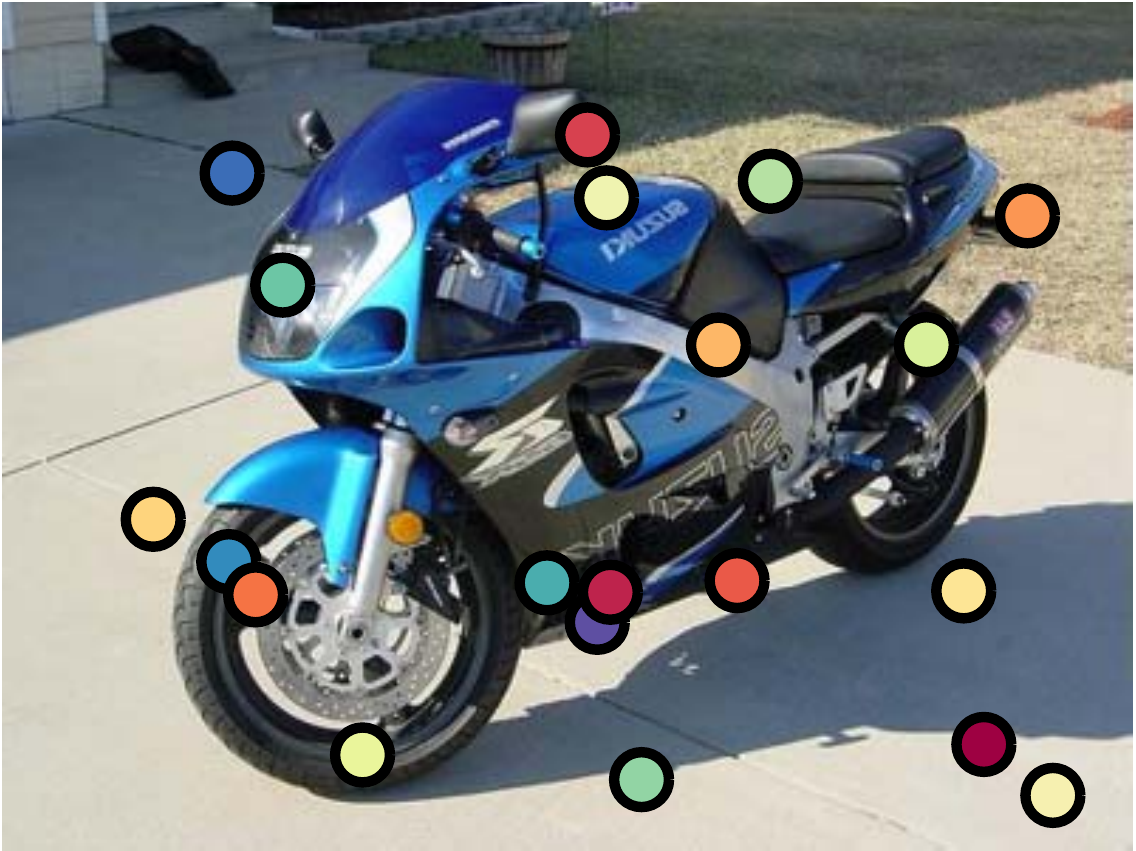}}\hfill
		\subfigure[]
		{\includegraphics[width=0.122\linewidth]{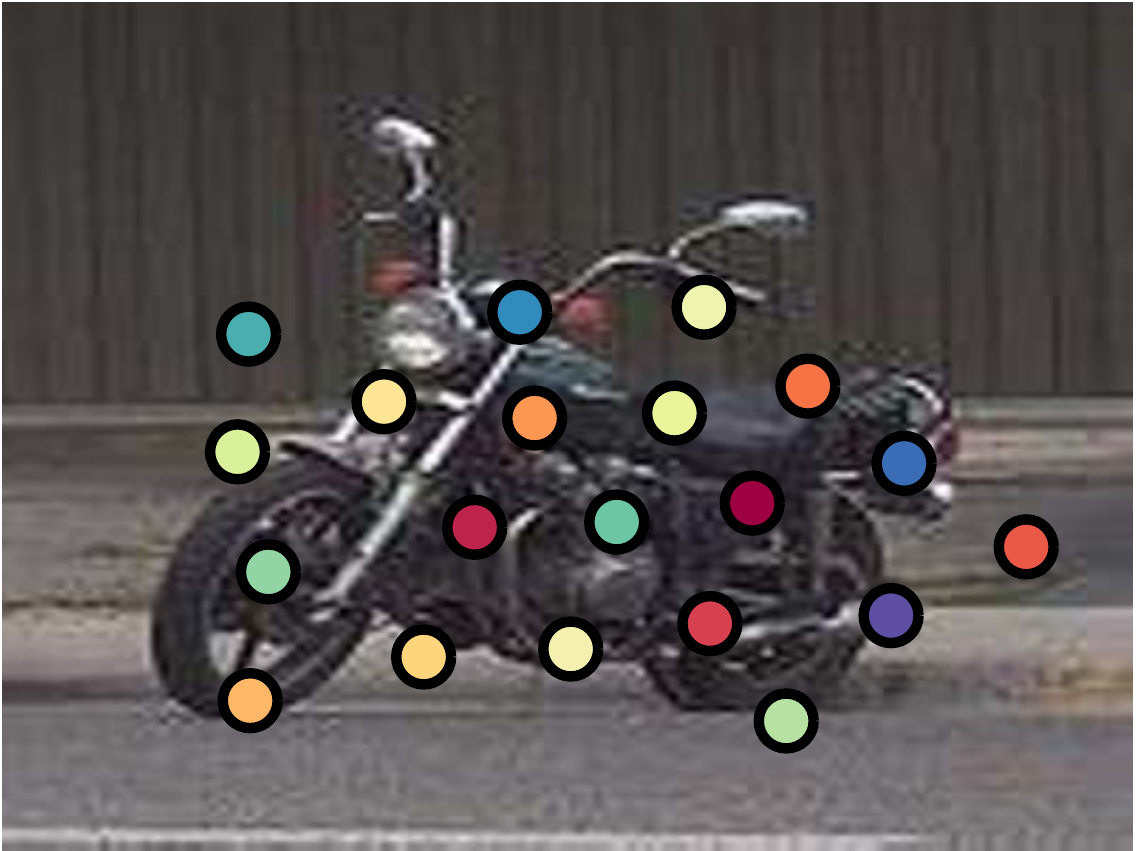}}\hfill
		\subfigure[]
		{\includegraphics[width=0.122\linewidth]{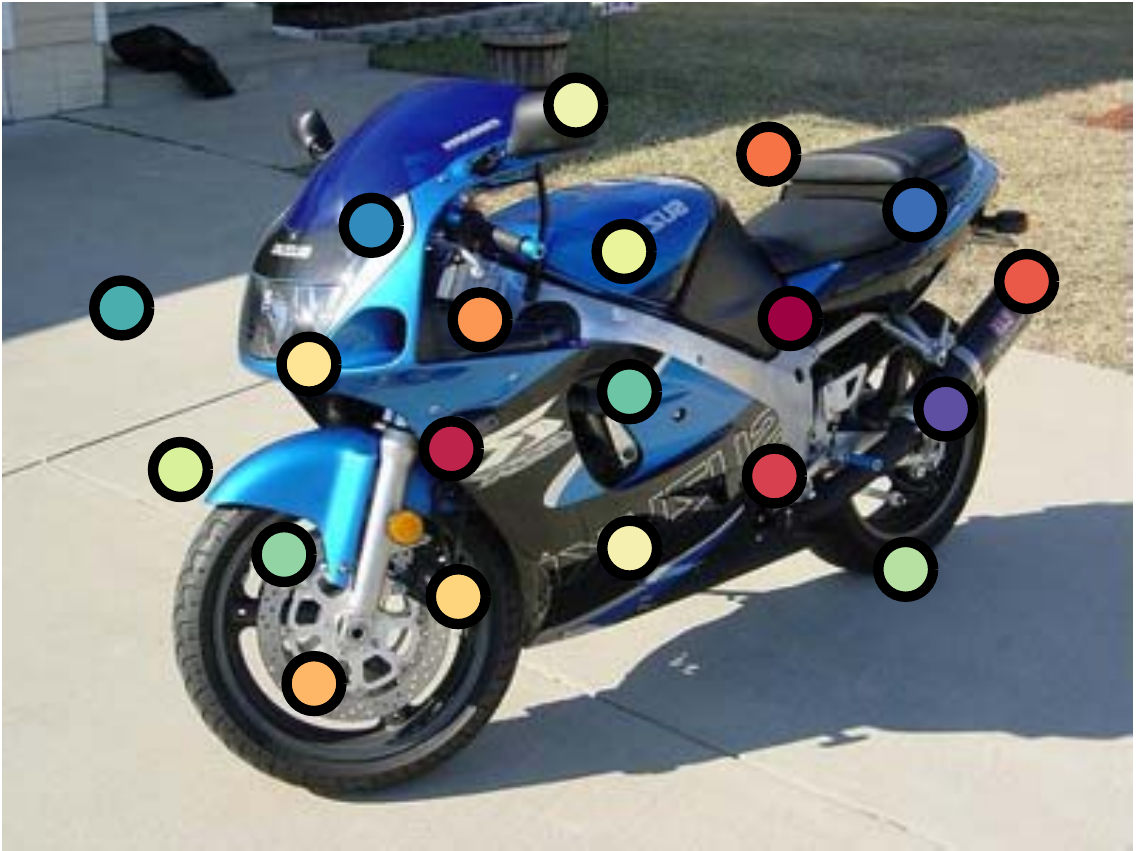}}\hfill
		\subfigure[]
		{\includegraphics[width=0.122\linewidth]{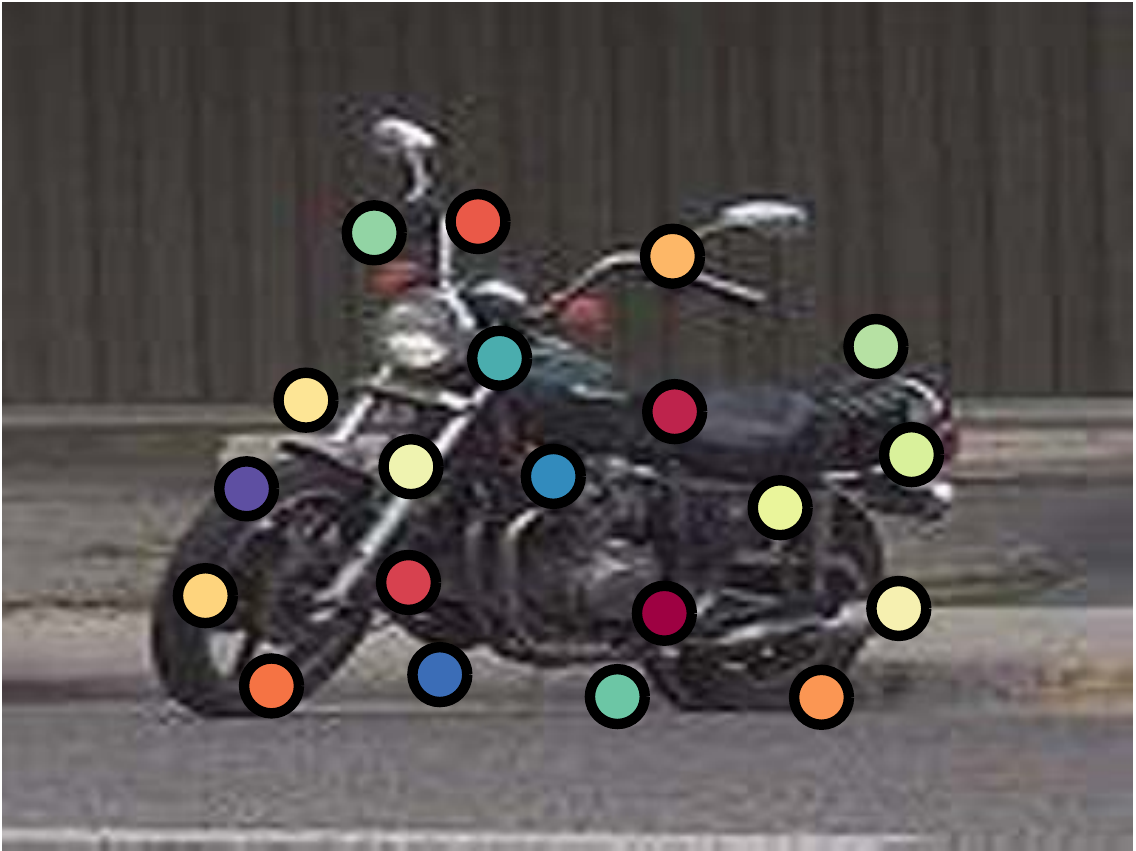}}\hfill
		\subfigure[]
		{\includegraphics[width=0.122\linewidth]{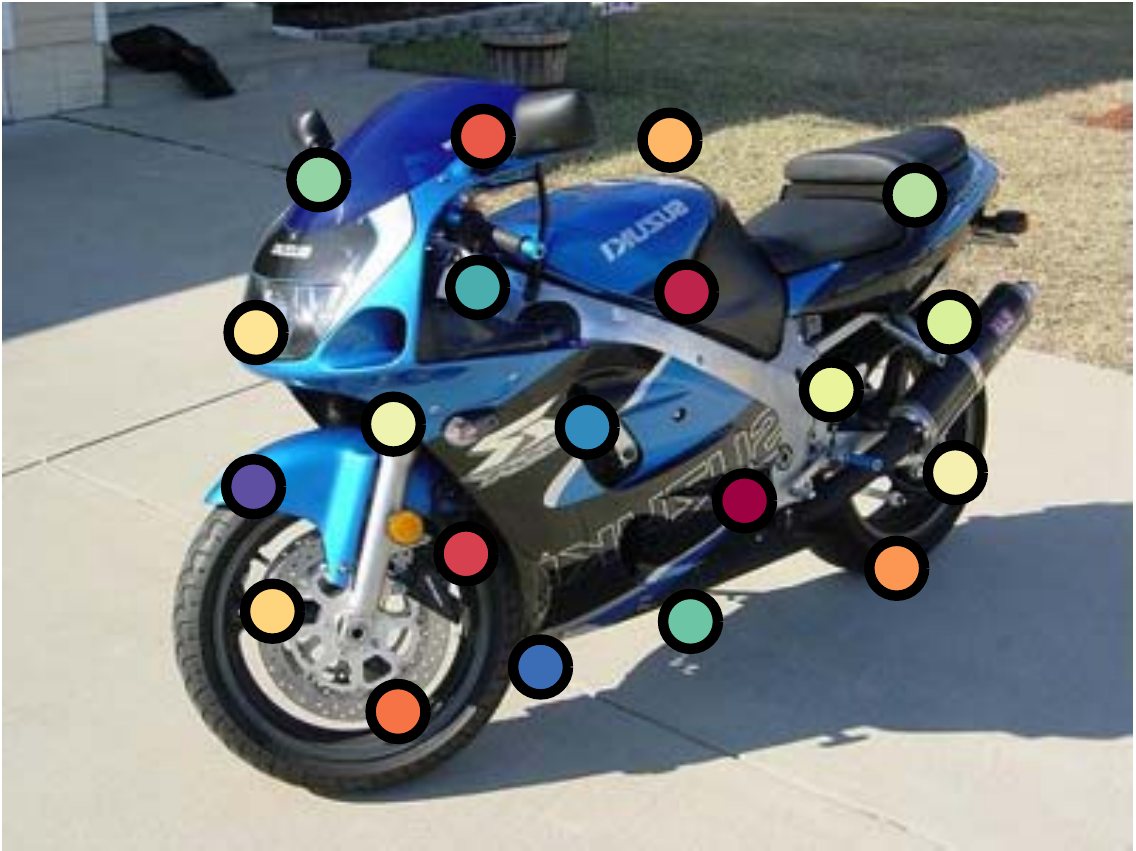}}\hfill
		\vspace{-21.5pt}
		\subfigure[(a)]
		{\includegraphics[width=0.122\linewidth]{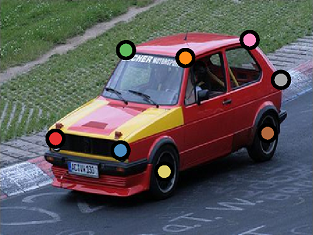}}\hfill
		\subfigure[(b)]
		{\includegraphics[width=0.122\linewidth]{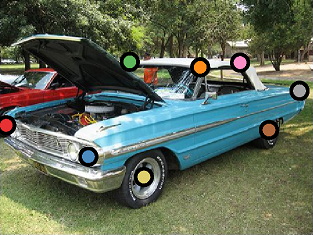}}\hfill
		\subfigure[(c)]
		{\includegraphics[width=0.122\linewidth]{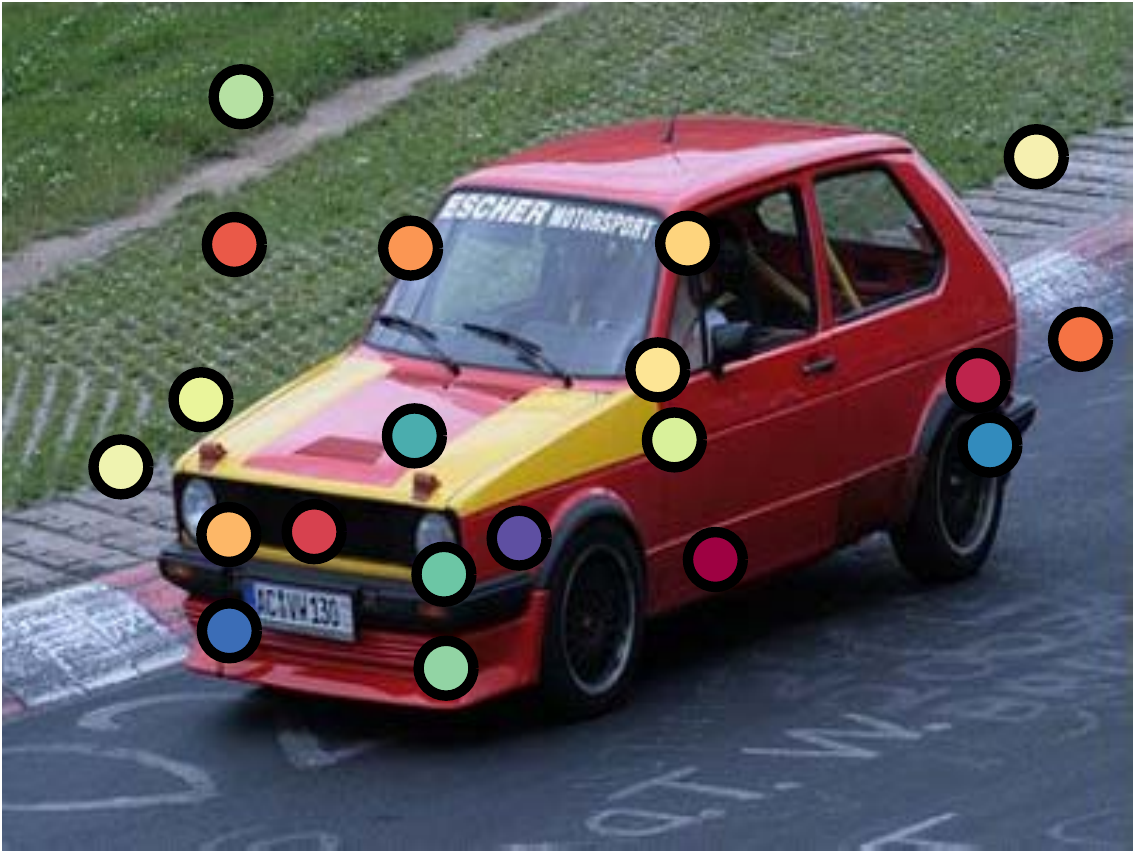}}\hfill
		\subfigure[(d)]
		{\includegraphics[width=0.122\linewidth]{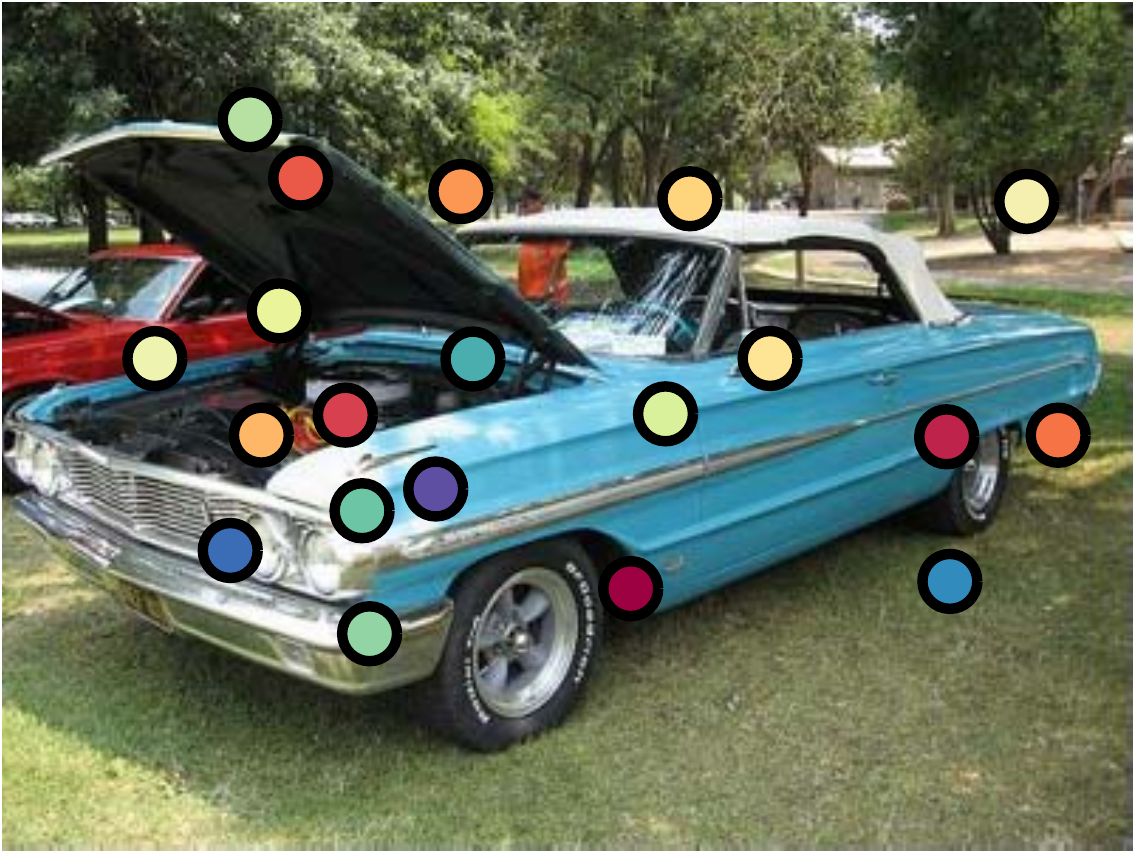}}\hfill
		\subfigure[(e)]
		{\includegraphics[width=0.122\linewidth]{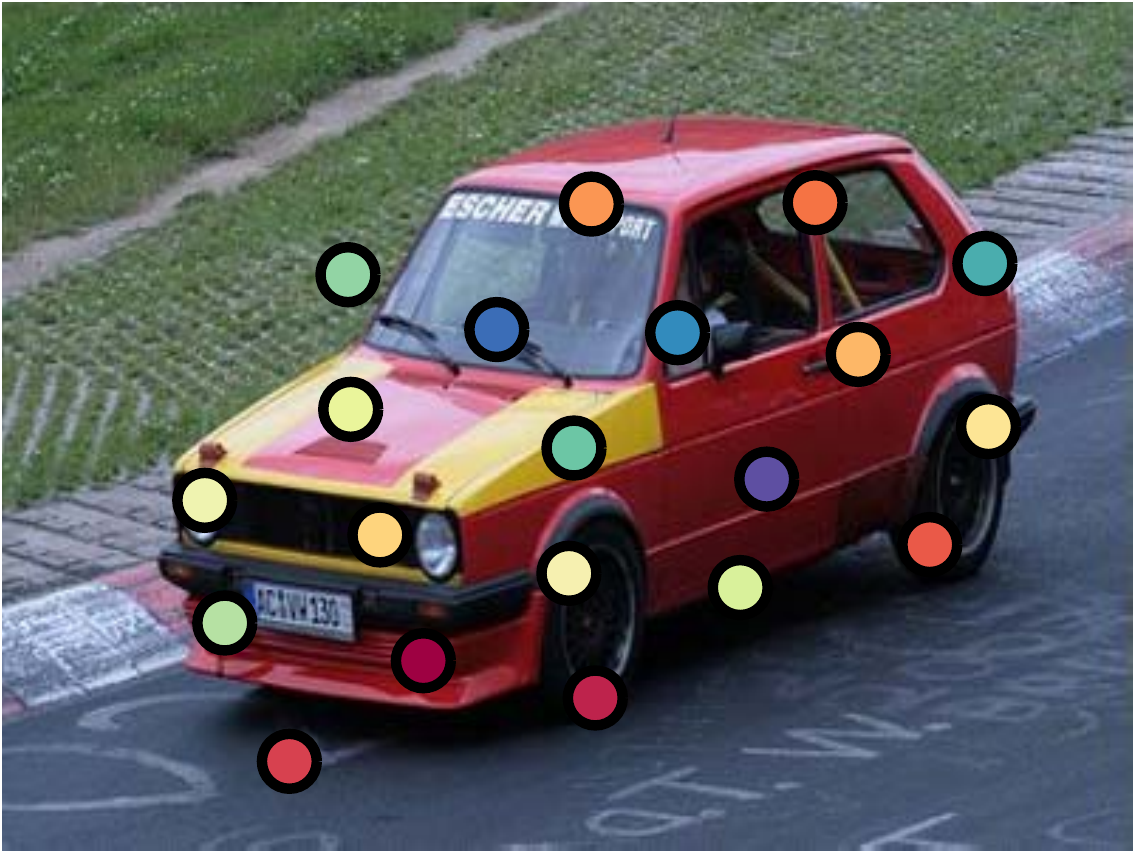}}\hfill
		\subfigure[(f)]
		{\includegraphics[width=0.122\linewidth]{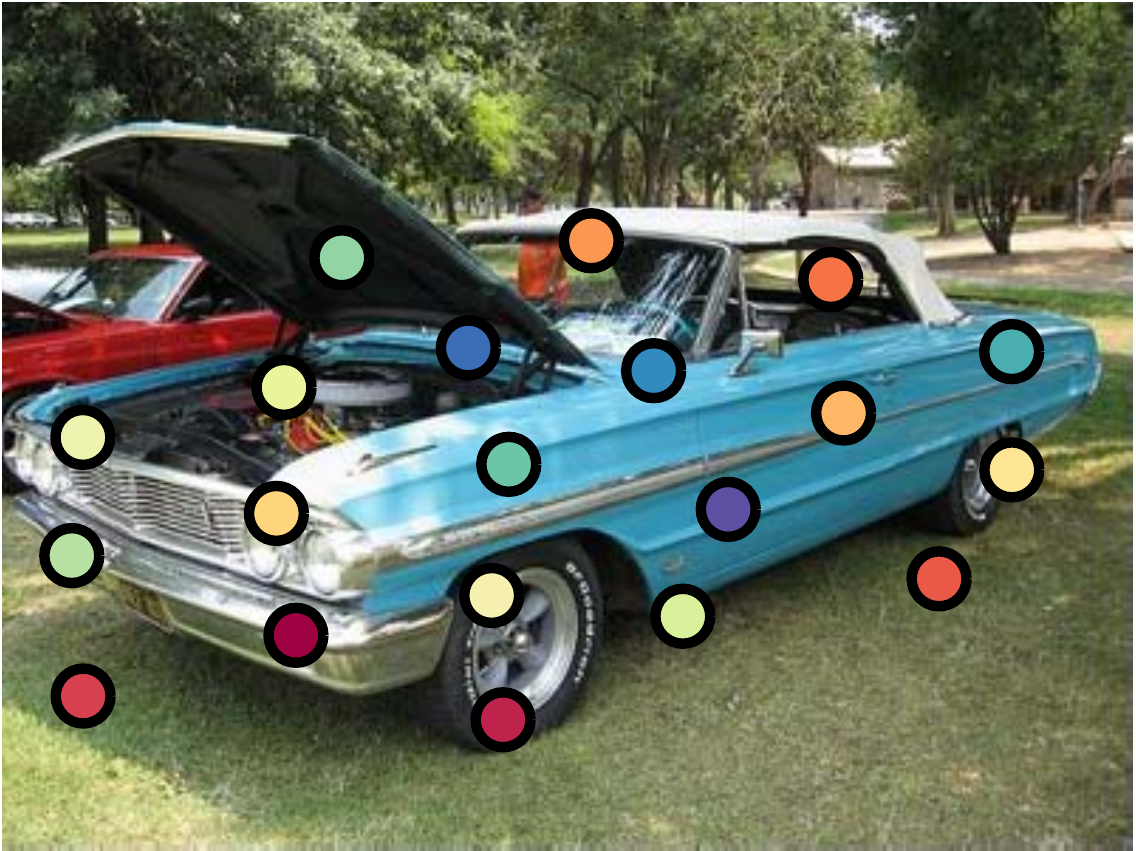}}\hfill
		\subfigure[(g)]
		{\includegraphics[width=0.122\linewidth]{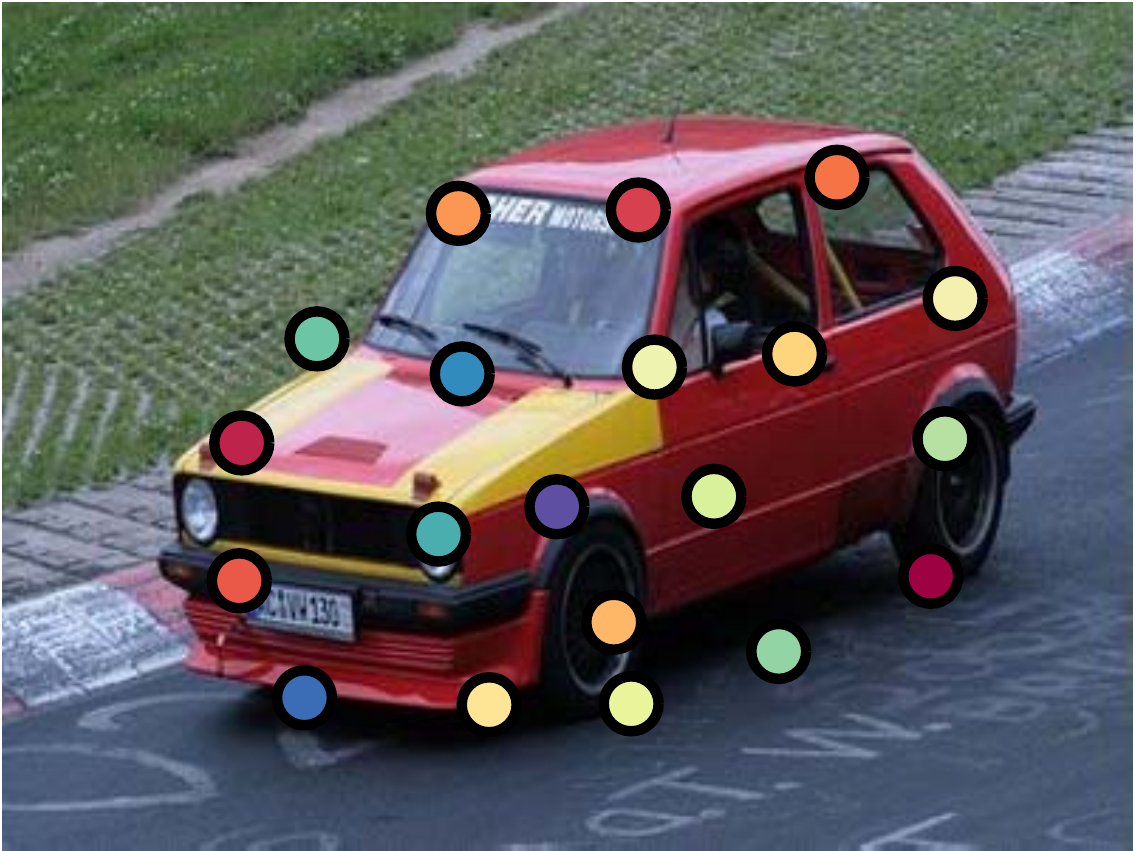}}\hfill
		\subfigure[(h)]
		{\includegraphics[width=0.122\linewidth]{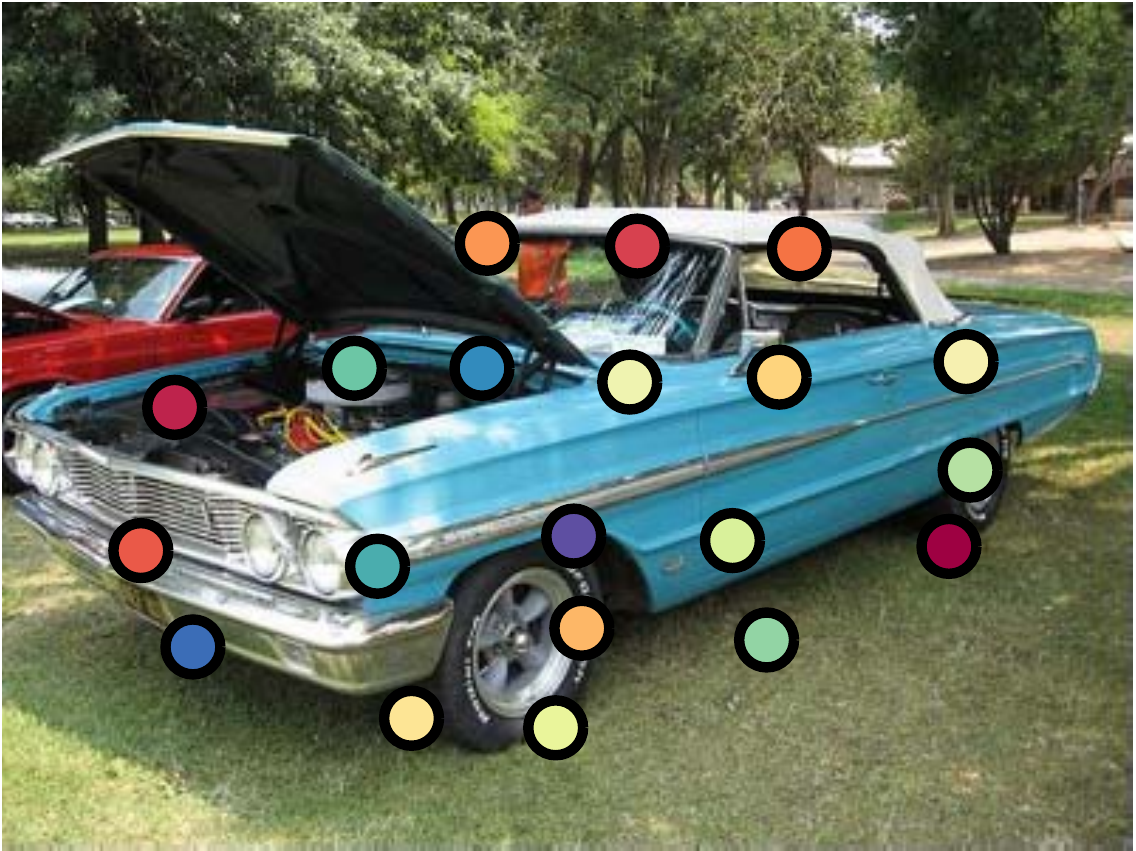}}\hfill
		\vspace{-3pt}
		\caption{Qualitative results of the object landmark detection on the JLAD dataset: (a), (b) ground-truth landmarks, the image pairs of~\figref{img:7} are used to discover landmarks with (c), (d) CIG~\cite{jakab18}, (e), (f) StrucRep~\cite{zhang18}, and (g), (h) Ours.}\label{img:8}\vspace{-10pt}
	\end{figure*}
	
	\paragraph{Semantic alignment}
	We evaluated our semantic alignment networks over 12 object categories on the JLAD dataset and the PF-PASCAL benchmark~\cite{Ham17}.
	For the evaluation metric, we used the percentage of correct keypoints (PCK) metric~\cite{yang2011articulated} which counts the number of keypoints having a transfer error below a given threshold $\alpha$, following the procedure employed in~\cite{han17}. 
	\tabref{tab:2} and~\tabref{tab:3} summarize the PCK values, and \figref{img:7} shows qualitative results.
	The results of detected landmarks of each image pair in~\figref{img:7} are visualized in~\figref{img:8}.
	As shown in~\tabref{tab:2},~\tabref{tab:3},~\figref{img:7} our results have shown highly improved performance qualitatively and quantitatively compared to the methods~\cite{rocco17,rocco18,jeon18,paul18} {that rely on synthetically or heuristically collected correspondence samples.
	This reveals the effect of the proposed joint learning technique where the structural smoothness is naturally imposed with respect to the detected object landmarks.
	This is in contrast to the methods that employ weak implicit smoothness constraints, such as image-level global transformation model~\cite{rocco17,rocco18,paul18}, locally constrained transformation candidates~\cite{kim18nips}, or local neighbourhood consensus~\cite{rocco18nips}.
	\vspace{-10pt}
	
	\begin{table}[t]
		\centering
		\begin{tabular}{
				>{\centering}m{0.30\linewidth} >{\centering}m{0.05\linewidth} >{\centering}m{0.10\linewidth} >{\centering}m{0.10\linewidth} 
				>{\centering}m{0.05\linewidth} >{\centering}m{0.15\linewidth}}
			\hlinewd{0.8pt}
			\multicolumn{1}{l|}{Methods} &$K$ &MAFL &ALFW &\multicolumn{1}{|c}{$K$} &JLAD\tabularnewline
			\hline
			\hline
			\multicolumn{1}{l|}{FPE~\cite{thewlis17iccv}} &50 &6.67 &10.53 &\multicolumn{1}{|c}{20} &13.32 \tabularnewline
			\multicolumn{1}{l|}{DEIL~\cite{thewlis17nips}} &- &5.83 &8.80 &\multicolumn{1}{|c}{-} &10.76 \tabularnewline
			\multicolumn{1}{l|}{StrucRep~\cite{zhang18}} &30 &3.16 &6.58 &\multicolumn{1}{|c}{20} &7.33 \tabularnewline
			\multicolumn{1}{l|}{CIG~\cite{jakab18}} &30 &3.08 &6.98 &\multicolumn{1}{|c}{20} &12.87 \tabularnewline
			\hline
			\multicolumn{1}{l|}{Ours wo/SS} &30 &3.58 &7.72 &\multicolumn{1}{|c}{20} &8.16 \tabularnewline
			\multicolumn{1}{l|}{\multirow{2}{*}{Ours}} &10 &3.33 &7.17 &\multicolumn{1}{|c}{10} &7.54 \tabularnewline
			\multicolumn{1}{l|}{} &30 &\bf{2.98} &\bf{6.51} &\multicolumn{1}{|c}{20} &\bf{6.92} \tabularnewline
			\hlinewd{0.8pt}
		\end{tabular}\vspace{+3pt}
		\caption{Comparison with state-of-the-art landmark detection techniques on the MAFL~\cite{zhang2014facial}, ALFW~\cite{koestinger2011annotated}, and JLAD dataset. $K$ denotes the number of used landmarks for linear regressor.}\label{tab:4}\vspace{-10pt}
	\end{table}
	
	\paragraph{Object landmark detection}
	We evaluated our landmark detection networks for human faces on MAFL and AFLW benchmarks~\cite{zhang2014facial,koestinger2011annotated}, including various objects on JLAD dataset.
	For the evaluation on MAFL benchmark~\cite{zhang2014facial}, we trained our model with facial image pairs in the CelebA training set excluding those appearing in the MAFL test set. For AFLW benckmark~\cite{koestinger2011annotated}, we further finetune the pretrained networks on AFLW training image sets, simlar to~\cite{zhang18,thewlis17iccv}.
	To evaluate our discovered landmarks’ quality, we use a linear model without a bias term to regress from the discovered landmarks to the
	human-annotated landmarks~\cite{zhang18,thewlis17iccv,thewlis17nips,jakab18}. Ground-truth landmark annotations of testing image pairs are provided to train this linear regressor.
	We follow the standard MSE metric in~\cite{zhang2014facial} and report performances in inter-ocular distance (IOD).
	\figref{img:8} shows qualitative results on JLAD dataset and \figref{img:9} for MAFL benchmark~\cite{zhang2014facial}.
	\tabref{tab:4} shows that our method achieves the state-of-the-art performance compared with existing models~\cite{zhang18,thewlis17iccv} that use synthesized image deformations for training their networks.
	The relatively modest gain on human faces compared to other object catogories may come from the limited appearance and geometric variations on MAFL and AFLW benchmarks, where the faces are cropped and aligned including little background clutters.
	A visual comparison of~\figref{img:8} and quantitative results of~\tabref{tab:4} demonstrate the benefits of joint learning with semantic alignment networks.
	Unlike existing methods~\cite{zhang18,thewlis17iccv,thewlis17nips,jakab18} that do not consider rich variations from the real image pairs, our method consistently discovers semantically meaningful landmarks over various object categories even under large appearance and shape variations.
	
	\begin{figure}[t]
		\centering
		\renewcommand{\thesubfigure}{}
		\subfigure[]
		{\includegraphics[width=0.195\linewidth]{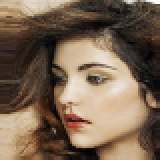}}\hfill
		\subfigure[]
		{\includegraphics[width=0.195\linewidth]{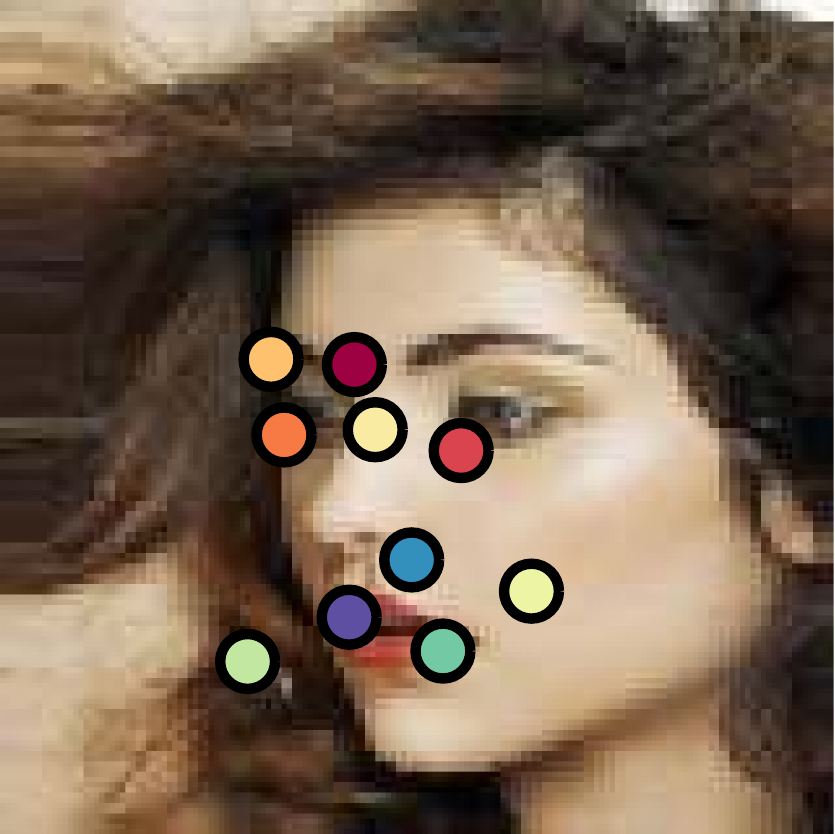}}\hfill
		\subfigure[]
		{\includegraphics[width=0.195\linewidth]{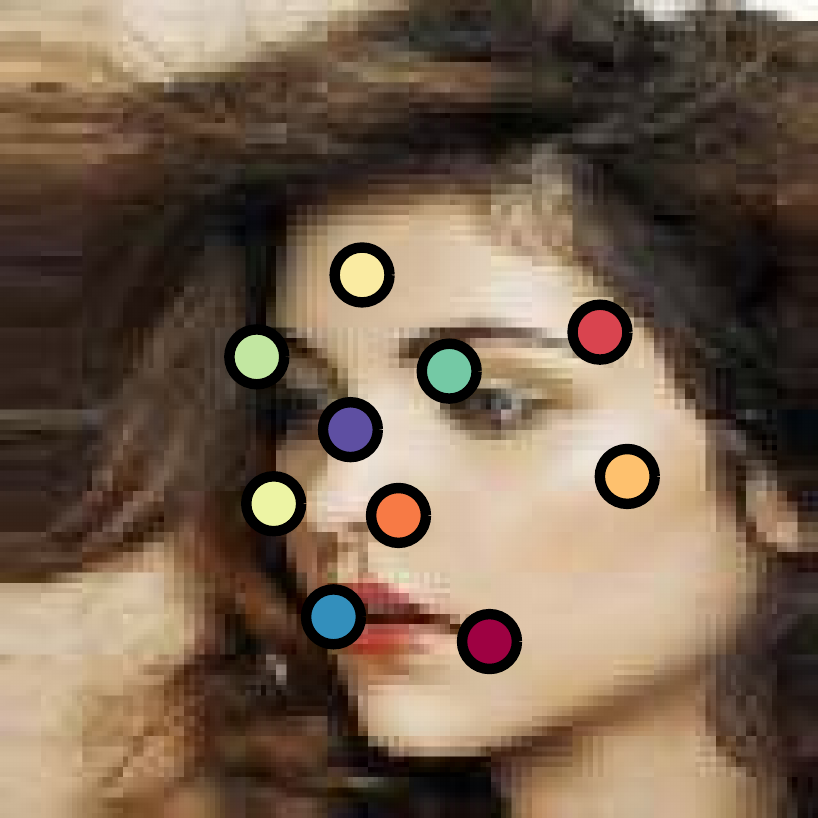}}\hfill
		\subfigure[]
		{\includegraphics[width=0.195\linewidth]{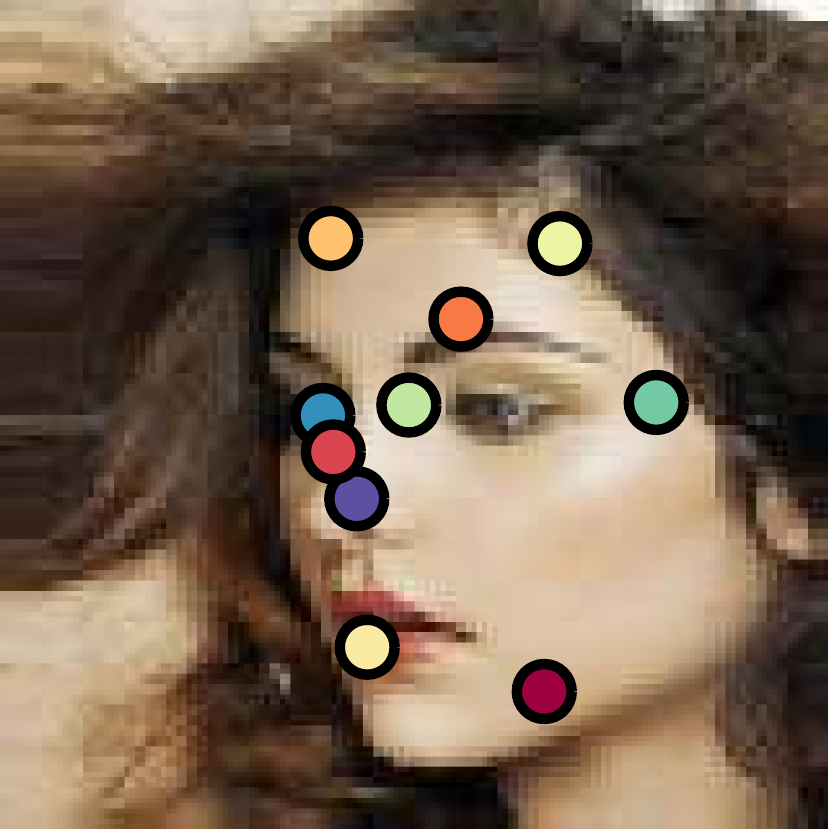}}\hfill
		\subfigure[]
		{\includegraphics[width=0.195\linewidth]{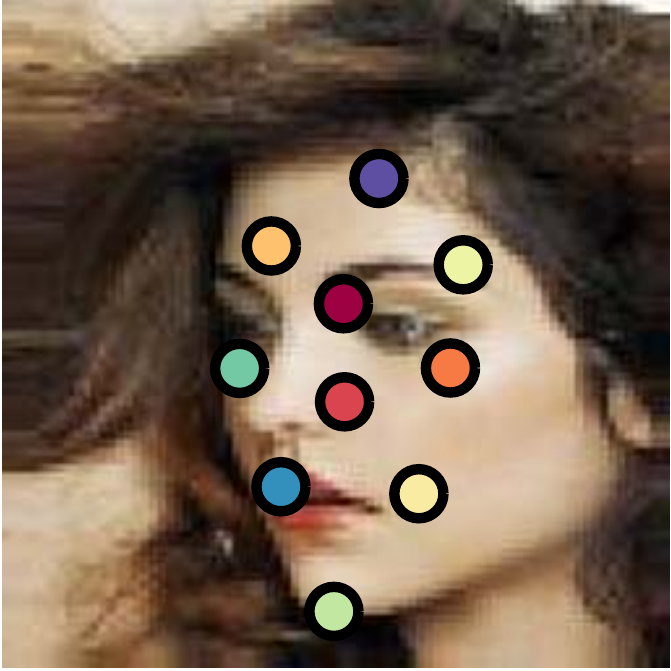}}\hfill
		\vspace{-22pt}
		\subfigure[(a)]
		{\includegraphics[width=0.195\linewidth]{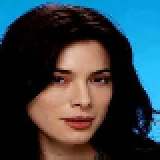}}\hfill
		\subfigure[(b)]
		{\includegraphics[width=0.195\linewidth]{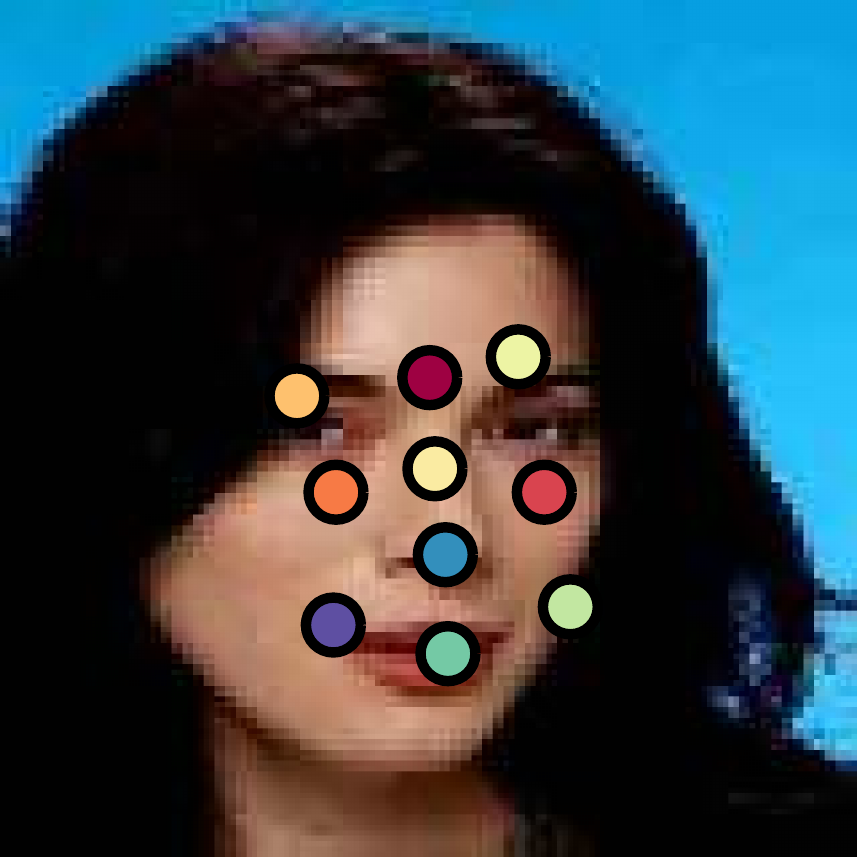}}\hfill
		\subfigure[(c)]
		{\includegraphics[width=0.195\linewidth]{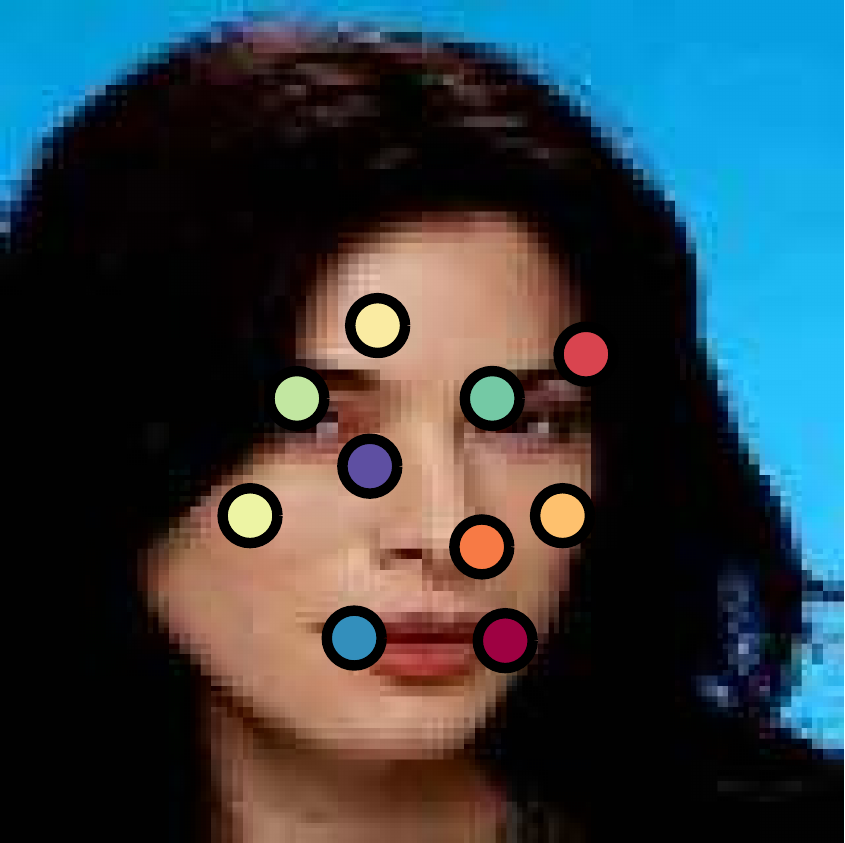}}\hfill
		\subfigure[(d)]
		{\includegraphics[width=0.195\linewidth]{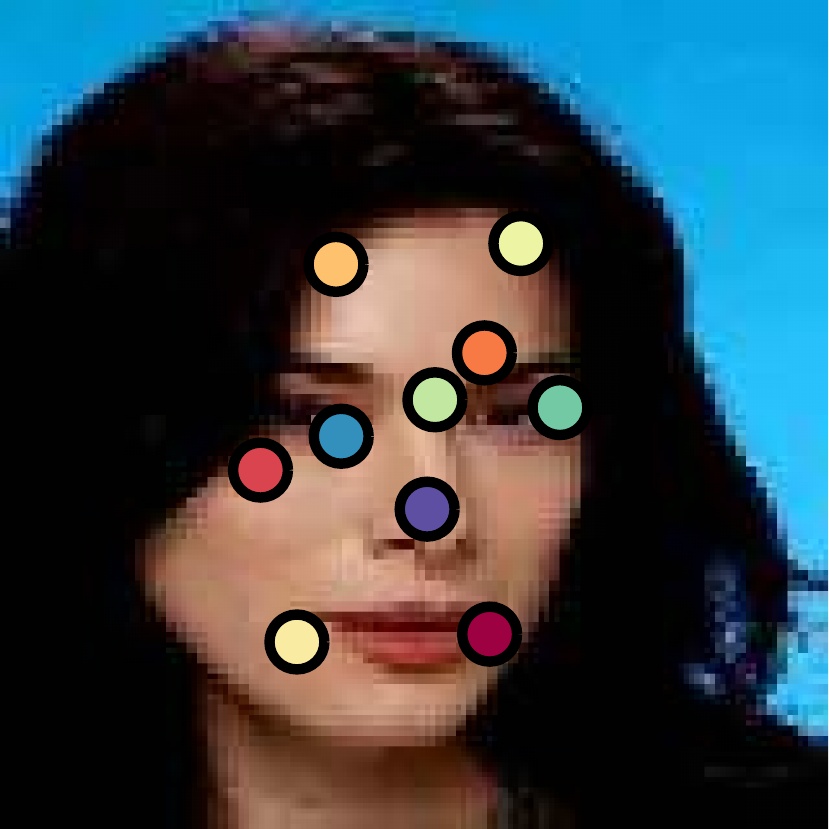}}\hfill
		\subfigure[(e)]
		{\includegraphics[width=0.195\linewidth]{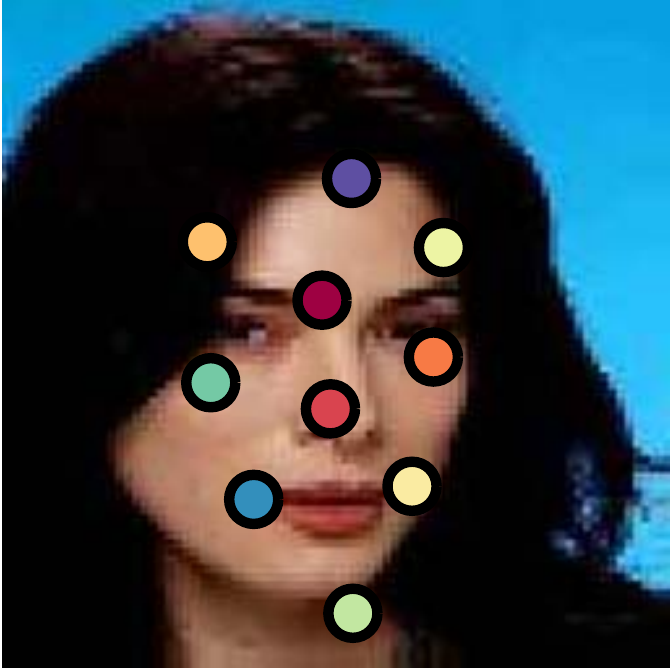}}\hfill
		\vspace{-3pt}
		\caption{Qualitative results of the object landmark detection on the MAFL benchmark~\cite{zhang2014facial}: (a) ground-truth landmarks, (b) FPE~\cite{thewlis17iccv}, (c) StrucRep~\cite{zhang18}, (d) CIG~\cite{jakab18}, (e) Ours.}\label{img:9}\vspace{-10pt}
	\end{figure}
	
	\begin{figure}[t]
		\centering
		\renewcommand{\thesubfigure}{}
		\subfigure[]
		{\includegraphics[width=0.195\linewidth]{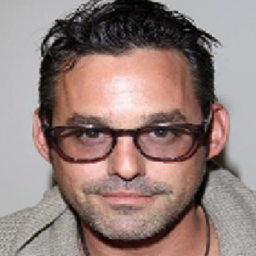}}\hfill
		\subfigure[]
		{\includegraphics[width=0.195\linewidth]{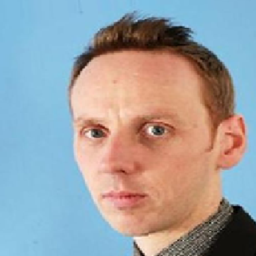}}\hfill
		\subfigure[]
		{\includegraphics[width=0.195\linewidth]{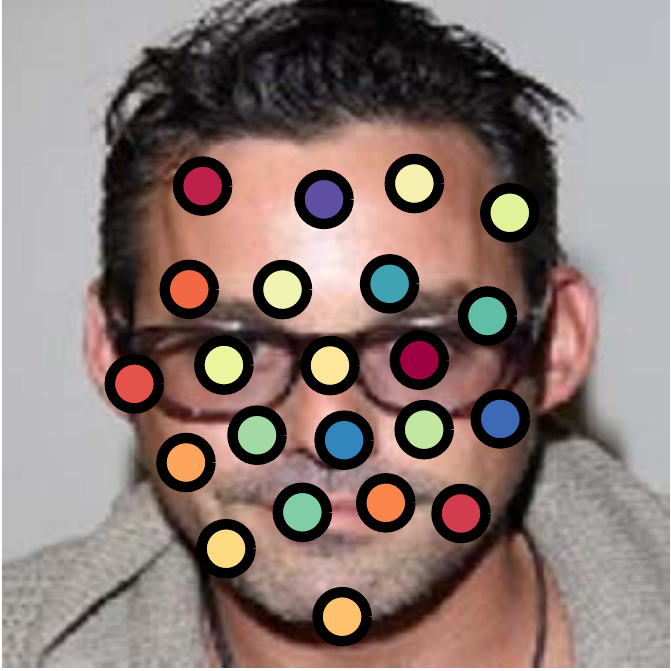}}\hfill
		\subfigure[]
		{\includegraphics[width=0.195\linewidth]{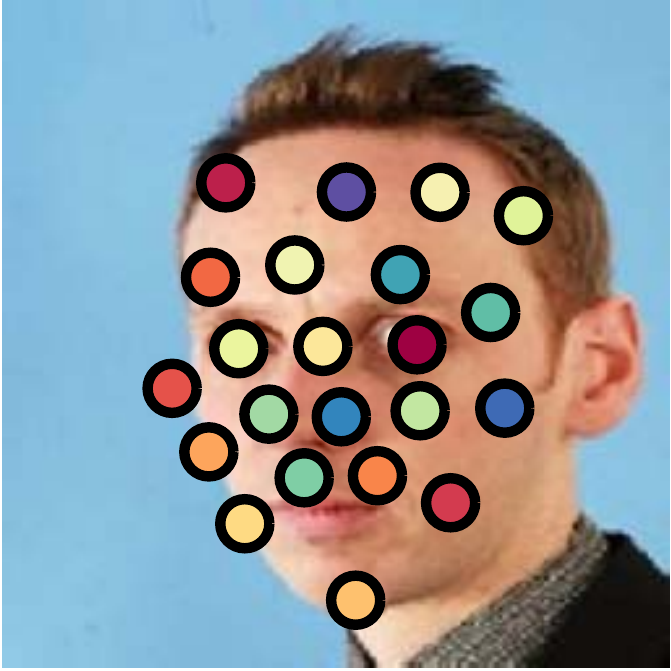}}\hfill
		\subfigure[]
		{\includegraphics[width=0.195\linewidth]{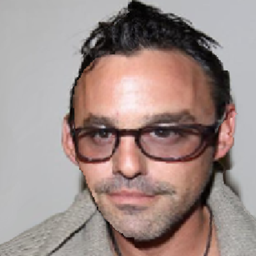}}\hfill
		\vspace{-22pt}
		\subfigure[(a)]
		{\includegraphics[width=0.195\linewidth]{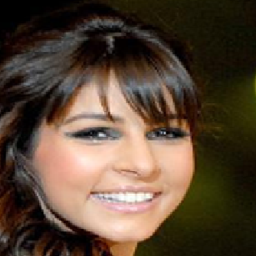}}\hfill
		\subfigure[(b)]
		{\includegraphics[width=0.195\linewidth]{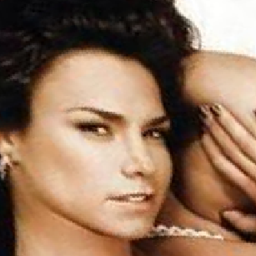}}\hfill
		\subfigure[(c)]
		{\includegraphics[width=0.195\linewidth]{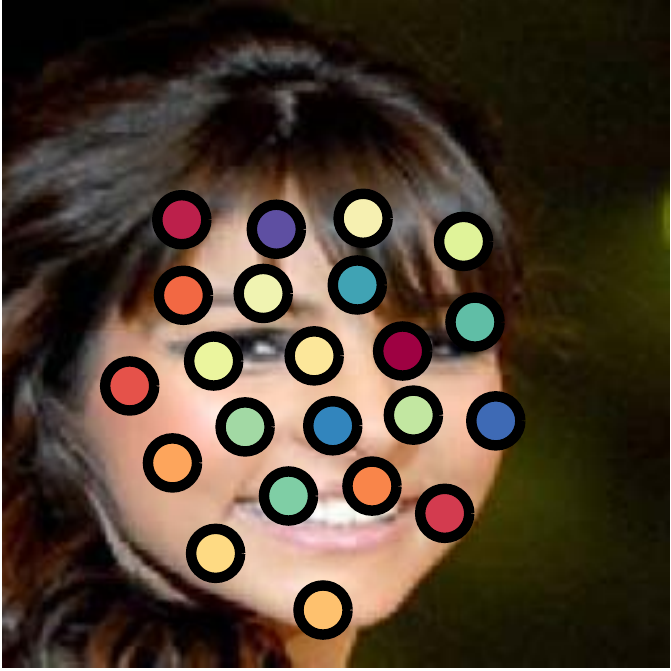}}\hfill
		\subfigure[(d)]
		{\includegraphics[width=0.195\linewidth]{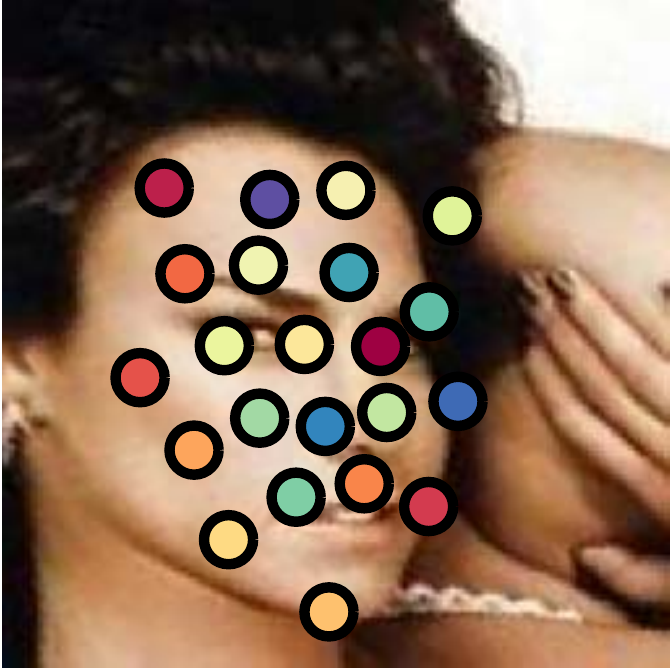}}\hfill
		\subfigure[(e)]
		{\includegraphics[width=0.195\linewidth]{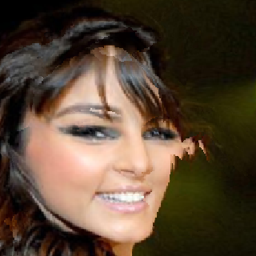}}\hfill
		\vspace{-3pt}
		\caption{Qualitative results of our semantic aligment networks on the MAFL benchmark: (a) source image, (b) target image,
			(c), (d) detected landmarks on source and target image, (e) warped image using correspondences.}\label{img:10}\vspace{-10pt}
	\end{figure}
	
	\section{Conclusion}
	We presented a joint learning framework for the landmark detection and semantic correspondence that utilizes the complementary interactions between the two tasks to overcome the lack of training data by alternatively imposing the consistent constraints.
	Experimental results on various benchmarks, including a newly introduced JLAD dataset, demonstrate the effectiveness of our method, such that the image pairs can be precisely aligned with the intrinsic structures of detected landmarks, and
	at the same time the landmarks can be consistently discovered with estimated semantic correspondence fields.

	{\small
		\bibliographystyle{ieee_fullname}
		\bibliography{egbib}
	}
	
\end{document}